\documentclass[11pt,a4paper]{article}
\usepackage[hyperref]{emnlp-ijcnlp-2019}
\usepackage{times}
\usepackage{latexsym}
\usepackage{amsmath}
\usepackage{amsfonts}
\usepackage{amsthm}
\usepackage{algorithmic}
\usepackage[ruled,vlined]{algorithm2e}
\usepackage{xcolor}
\usepackage{graphicx}
\usepackage{placeins}
\theoremstyle{plain}
\newtheorem{theorem}{Theorem}
\newtheorem{definition}[theorem]{Definition}

\newtheorem{lemma}[theorem]{Lemma}

\newcommand{\argmin}{\operatornamewithlimits{\arg \min}}

\newcommand{\x}{\boldsymbol{x}}
\newcommand{\X}{\mathcal{X}}

\newcommand{\R}{\mathbb{R}}
\newcommand{\E}{\mathbb{E}}
\newcommand{\Epos}{\mathbb{E}_{\mathrm{P}}}
\newcommand{\Eneg}{\mathbb{E}_{\mathrm{N}}}

\newcommand{\zerooneloss}{\ell_{0\text{-}1}}

\newcommand{\sell}{\ell_\mathrm{sym}}

\usepackage{booktabs} 
\usepackage{multirow}
\usepackage{url}

\usepackage{array}
\usepackage{enumitem}
\newcolumntype{L}{>{$}l<{$}}
\newcolumntype{C}{>{$}c<{$}}

\newcommand{\corrrisk}[1]{{R^{\sell}_{\mathrm{AUC}\text{-}\mathrm{Corr}}(#1) }}
\newcommand{\corremprisk}[1]{{\widehat{R}^{\sell}_{\mathrm{AUC}\text{-}\mathrm{Corr}}(#1) }}

\aclfinalcopy % Uncomment this line for the final submission

\title{Learning Only from Relevant Keywords and Unlabeled Documents}

\author{Nontawat Charoenphakdee\textsuperscript{1,3} \qquad
  Jongyeong Lee\textsuperscript{1,3} \qquad 
  Yiping Jin\textsuperscript{2} \\
  \textbf{ \qquad Dittaya Wanvarie\textsuperscript{2} \qquad
  Masashi Sugiyama\textsuperscript{3,4,1}} \\ 
  \textsuperscript{1}Department of Computer Science, The University of Tokyo \\
  \textsuperscript{2}Department of Mathematics and Computer Science, Faculty of Science, Chulalongkorn University\\
  \textsuperscript{3}RIKEN Center for Advanced Intelligence Project \\
  \textsuperscript{4}Department of Complexity Science and Engineering, The University of Tokyo\\
  {\tt \{nontawat, lee\}@ms.k.u-tokyo.ac.jp} \\
    {\tt jinyiping@knorex.com}\\
  {\tt Dittaya.W@chula.ac.th}\\
    {\tt sugi@k.u-tokyo.ac.jp}
  }
\date{}

\begin{document}
\maketitle
\begin{abstract}
We consider a document classification problem where document labels are absent but only relevant keywords of a target class and unlabeled documents are given. Although heuristic methods based on pseudo-labeling have been considered, theoretical understanding of this problem has still been limited. Moreover, previous methods cannot easily incorporate well-developed techniques in supervised text classification. In this paper, we propose a theoretically guaranteed learning framework that is simple to implement and has flexible choices of models, e.g., linear models or neural networks.
We demonstrate how to optimize the area under the receiver operating characteristic curve (AUC) effectively and also discuss how to adjust it to optimize other well-known
evaluation metrics such as the accuracy and $\mathrm{F}_{1}$-measure.
Finally, we show the effectiveness of our framework using benchmark datasets.
\end{abstract}

\section{Introduction}
Supervised text classification is a traditional problem in natural language processing that has been studied extensively~\citep{nigam2000text,sebastiani2002machine,forman2003extensive,joulin2016bag}.
In this problem, we are given a set of labeled text documents and the goal is to construct a classifier that can classify an unseen document effectively.
There are many useful applications for text classification, e.g., sentiment classification~\citep{liu2012survey,medhat2014sentiment}, 
biomedical text mining~\citep{cohen2005survey, huang2015community}, and social media monitoring~\citep{zeng2010social,hu2012text}. 

In the real-world, it is impractical to expect that labeled data can always be obtained abundantly.
For example, given big data of unlabeled texts, it can be very time-consuming and costly for the labeling task so that we can apply a supervised text classification method.
Another example is when the label information is protected due to privacy concerns.
These bottlenecks motivate researches on weakly-supervised learning~\citep{zhou2017brief}, which focuses on devising a machine learning method that can learn effectively although a lot of labeled data are not accessible.

An alternative is to provide keywords as a hint for classifying a document to a target class. 
Intuitively, this approach can be much cheaper when the number of documents is huge since the number of keywords does not necessarily grow linearly with the number of documents. 
\emph{Dataless classification} is a text classification problem where we are given keywords for each class and unlabeled documents~\citep{chang2008importance,song2014dataless,chen2015dataless,li2018pseudo}. 
Note that we do not have access to \emph{any} labeled documents.

In this paper, we investigate dataless classification in the situation where only the keywords of the target class are given. 
We regard documents of a class that we have relevant keywords as positive and others as negative.
Unlike dataless classification, we have access to only the keywords of the positive class. %but we have no information about the keywords of a negative class. 
This problem setting can be considered more difficult than the traditional one since one can always reduce dataless classification to this problem by ignoring negative keywords.
This scenario is also highly relevant when the information about negative classes cannot be explicitly described, e.g., in the information retrieval task, we may only have information about the target class of interest.

This problem setting has already been considered by~\citet{jin2017combining}, where they called it \emph{lightly-supervised one-class classification}.
To the best of our knowledge, existing work in dataless classification and lightly-supervised one-class classification neither provide a theoretical guarantee, nor have flexible choices of models and optimization algorithms for this problem.

The goal of this paper is to formalize lightly-supervised one-class classification and develop a reliable and flexible framework to handle this problem effectively.
To achieve this goal, we propose a framework that has a theoretical guarantee and allows practitioners to have flexible choices to maximize the performance, e.g., neural network architectures such as convolutional neural networks~\citep{zhang2015character} or recurrent neural networks~\citep{lai2015recurrent}. They can also pick an optimization method that is suitable for their model such as Adam~\citep{adam} or AMSGrad~\citep{adam_amsgrad}. These advantages allow us to utilize well-developed supervised-learning text classification methods in our framework.

We point out that the problem of lightly-supervised one-class classification is highly related to binary classification from corrupted labels. Intuitively, the common idea to solve this problem is to use keywords to pseudo-label given documents then performs classification~\citep{jin2017combining}. 
However, pseudo-positive and pseudo-negative labels are unreliable since they can be incorrectly labeled. 
Thus, the learning method should be robust against label corruption.
In our framework, we also use relevant keywords to split unlabeled data into two sets. 
Then, the key idea is to employ a method based on an empirical risk minimization framework~\citep{vapnik1998statistical} with a loss function that benefits from a symmetric condition to maximize AUC~\citep{charoenphakdee2019symmetric}.
We justify the soundness of our method by proving that our proposed framework gives a consistent estimation of the true expected AUC risk as long as the divided two sets have  different proportions of positive data. 
%This minimal condition is not difficult to satisfy as shown in experiments.
We elucidate this theoretical result by providing an estimation error bound for AUC maximization using a Rademacher complexity measure. 
Furthermore, we discuss how to adjust our method to optimize other evaluation metrics, which are the accuracy and $\mathrm{F}_{1}$-measure.
We also illustrate that label corruption may cause a classifier to pick a wrong decision boundary and suggest to adjust the threshold on the basis of the proportion of positive data in unlabeled data. 

\section{Preliminaries}
In this section, we introduce the notation, review AUC, and an empirical risk minimization framework for optimizing AUC.
\subsection{Notation}
Let $\x \in \X$ be a pattern in the input space $\X$ and $y \in \{-1, 1\}$ be a label. We denote $g:\X \to \R$ as a prediction function. 
Let $\Epos[\cdot] $ and $\Eneg[\cdot]$ denote the expectation over the class-conditional probability $p(\x|y=1)$ and $p(\x|y=-1)$, respectively.
Moreover, let $\pi$ denote a class prior $p(y=1)$, i.e., the proportion of positive data. 

Loss functions that we will discuss in this paper are in the family of margin losses, where a loss receives one argument~$\ell: \R \to \R$. 
This loss family covers a lot of well-known losses~\citep{bartlett2006}, e.g., the logistic loss, squared loss, and hinge loss. 
We define the zero-one loss as $\zerooneloss (z) = -\frac{1}{2} \mathrm{sign}(z) + \frac{1}{2},$
where sign$(g(\x))=1$ if $g(\x)>0$, $-1$ if $g(\x)<0$, and $0$ otherwise.
Finally, we consider a symmetric loss as a margin loss $\sell$ that satisfies the symmetric condition $\sell(z)+\sell(-z)=K$, where $K$ is a positive constant. 
Examples of such losses are the zero-one loss and sigmoid loss~\citep{ghosh2015making}. Table~\ref{tab:loss-functions} provides examples of margin losses in binary classification.

\begin{table}[t]
\centering
\caption{Examples of loss functions. A symmetric loss is a loss function such that $\ell(z)+\ell(-z) = K$, where $K$ is a positive constant.}
\label{tab:loss-functions}
\begin{tabular}{CCC}
\hline
\text{Loss name} & \ell(z)  & \text{Symmetric} \\ \hline
%\text{Ramp} & -0.5 \max(0, \min(1, 0.5-0.5z)) &\times\\ 
% \text{Hinge} & \max(0, 1-z)  &\times \\ 
\text{Logistic} & \log(1+e^{-z}) &\times \\ 
\text{Squared} & (1-z)^{2}  &\times\\ 
\text{Zero-one} & -\frac{1}{2} \mathrm{sign}(z) + \frac{1}{2} &\checkmark \\ 
% \text{Unhinged} & 1-z &\checkmark \\ 
\text{Sigmoid} & \frac{1}{1+\exp(z)} &\checkmark\\ \hline
\end{tabular}
\end{table}

\begin{figure*}
\vspace{0.1in}
\begin{center}
\centerline{\includegraphics[scale =0.5]{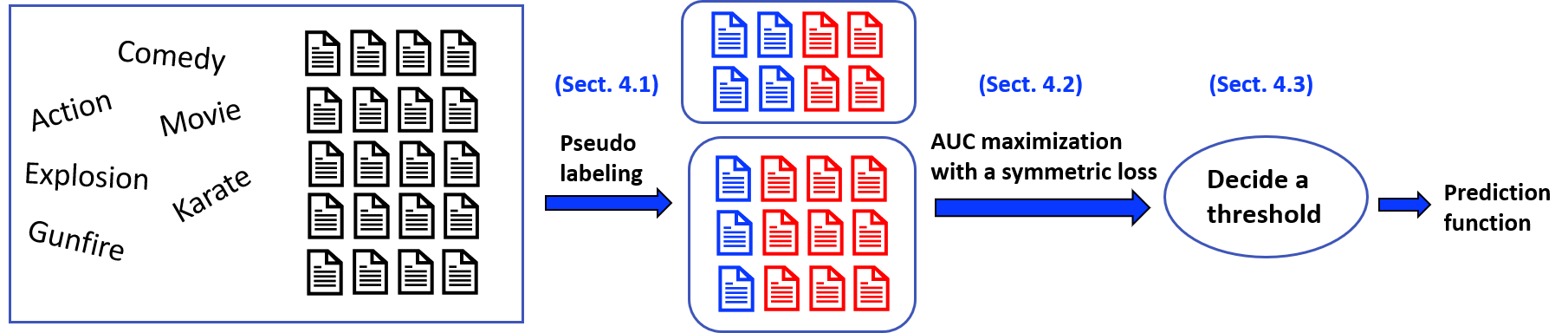}}
\caption{An overview of the framework. Blue documents indicates clean positive data and red documents denote clean negative data in the two sets of data divided by the pseudo-labeling algorithm. Note that labels are not observed by the framework. Sect. denotes a section that describes the procedure for each step.}
\label{fig:ete-network}
\end{center}
\vspace{-0.35in}
\end{figure*}

\subsection{Area under the Receiver Operating Characteristic Curve (AUC)}
AUC is an evaluation metric for a bipartite ranking task~\citep{cortes2004auc, menon2016bipartite}, where we want to find a function that outputs a higher value for positive data than negative data. 
Moreover, AUC is also a popular metric for a classifier under class imbalance~\citep{menon2015}. It is also known in the literature of statistics as the Wilcoxon-Mann-Whitney
statistic~\citep{mann1947test,hanley1982meaning}.

Let us consider the AUC risk, i.e., \emph{bipartite ranking risk}~\citep{narasimhan2013relationship}:
\begin{align}\label{pnrank}
R_{\mathrm{AUC}}^{\zerooneloss}(g) = \Epos \left[ \Eneg \left[ \zerooneloss(g(\x^{\mathrm{P}}_i)-g(\x^{\mathrm{N}}_j)) \right] \right]\text{.}
\end{align}
Then, AUC is defined as $\mathrm{AUC}(g) = 1-R_{\mathrm{AUC}}^{\zerooneloss}(g)$.
It is important to note that unlike many other evaluation metrics, we can evaluate AUC without deciding the threshold for a function $g$.
In this paper, we will show that optimizing AUC in our problem setting has a great advantage when no information about the threshold is given since we only have keywords and unlabeled documents. 
\subsection{Empirical Risk Minimization (ERM) for AUC Maximization}
Here, we review a widely used framework in machine learning called the ERM framework ~\citep{vapnik1998statistical} for AUC maximization. 

The goal of AUC maximization is to minimize the AUC risk in \eqref{pnrank} with respect to the zero-one loss $R_{\mathrm{AUC}}^{\zerooneloss}(g)$~\citep{menon2016bipartite}.

Although the goal is to minimize the bipartite ranking risk with respect to the zero-one loss $\zerooneloss$, a surrogate loss $\ell$, e.g., the logistic loss or the squared loss, is instead applied in practice since minimizing $\zerooneloss$ is computationally infeasible~\citep{zeroonenphard2,  zhang2004statistical, bartlett2006,zeroonenphard1}.
Note that we cannot directly minimize the AUC risk in our setting since we do not have access to the positive and negative distributions.

To explain the ERM framework for bipartite ranking, in this section, let us assume that we are given positive data $\{\x^{\mathrm{P}}_i\}_{i=1}^{n_\mathrm{P}}$ and negative data $\{\x^{\mathrm{N}}_j\}_{j=1}^{n_\mathrm{N}}$ drawn from the class-conditional probability densities $p(\x|y=1)$ and $p(\x|y=-1)$, respectively.
Having access to positive and negative data, the ERM framework suggests us to minimize the following empirical risk~\citep{yan2003optimizing,cortes2004auc}:
\begin{align}\label{pnrankemp}
\hat{R}_{\mathrm{AUC}}^\ell(g) = \frac{1}{n_\mathrm{P}n_\mathrm{N}}  \sum_{i=1}^{n_\mathrm{P}} \sum_{j=1}^{n_\mathrm{N}} \ell(g(\x^{\mathrm{P}}_i)-g(\x^{\mathrm{N}}_j))    \text{.}
\end{align}
With the given training data and a surrogate loss~$\ell$, we can minimize $\hat{R}_{\mathrm{AUC}}^\ell(g)$.
This risk estimator $\hat{R}_{\mathrm{AUC}}^\ell(g)$ is an unbiased and consistent estimator of the true AUC risk~\eqref{pnrank}, which means it converges to the true AUC risk as the number of data increases~\citep{yan2003optimizing,herschtal2004optimising,gao2015consistency}.

Note that in practice, it is common to apply a regularization method for controlling the bias-variance trade-off of the estimator to avoid overfitting.
In this paper, we adopt the ERM framework for minimizing the AUC risk.
Note that one challenge of our problem is that we cannot apply the standard AUC risk directly since we have no access to any labeled data. 

\section{Problem Formulation}
In this section, we formulate the problem of learning from keywords and unlabeled data. We are given a set of relevant keywords 
% \begin{align*}
$W:=  \{w_j\}_{j=1}^{n_\mathrm{W}}$,
% \end{align*}
and unlabeled documents drawn from the following distribution:
\begin{align*}
    X_{\mathrm{U}}:=  \{\x^\mathrm{U}_i\}_{i=1}^{n_\mathrm{U}}\stackrel{\mathrm{i.i.d.}}{\sim} p_{\pi}(\x),
\end{align*}
where 
\begin{align*}
    p_{\pi}(\x) = \pi p(\x|y=1) +(1-\pi)\, p(\x|y=-1).
\end{align*}
The evaluation metric defines the goal of this problem. 
For example, if we want a ranking function that outputs a large value for positive data, and small otherwise, the well-known evaluation metric is AUC~\citep{agarwal2005roc}. We will discuss other evaluation metrics, which are the accuracy and $\mathrm{F}_{1}$-measure in Section~\ref{sec:eva-metric}.
\section{Proposed Framework}

In this section, we propose a novel framework for handling this problem systematically. The framework consists of three parts. First, we use relevant keywords to extract pseudo-positive data from the unlabeled data. 
Then, we employ a symmetric loss function for AUC optimization within the empirical risk minimization framework. Finally, we adjust the threshold of a trained AUC maximizer to further optimize other evaluation metrics. Figure~\ref{fig:ete-network} shows an overview of our framework.

\subsection{Pseudo-labeling}
Here, we propose a simple pseudo-labeling method. We want to emphasize that we do not expect that a pseudo-labeling algorithm is perfect, i.e., pseudo-positive documents may contain negative documents. 
In Section~\ref{sect:AUC-sym}, we show that it is possible to maximize AUC effectively even though the labels are highly corrupted.

Since we focus on proposing a general framework that can be used in a wide range of text classification tasks, we use a simple pseudo-labeling method.
In practice, one can incorporate prior knowledge such as the specific characteristic of documents to develop a sophisticated and potentially better pseudo-labeling algorithm to further enhance the performance. 
Our simple pseudo-labeling method is a ranking score based on the document similarities with the cosine similarity  between unlabeled documents and keywords.
Although we assume some keywords are given, 
it is hard to guarantee that the given keywords are sufficient to learn effectively.
Hence, it is reasonable to sample more similar words to given keywords. 
In this paper, we used the GloVe model to find similar words \cite{pennington2014glove}.
With a sufficient number of keywords, we calculate the cosine similarity between the set of extended keywords set and each unlabeled document by simply merging all keywords and extended keywords into one document and measure the similarity between this keyword document and each unlabeled document. 
Finally, we pseudo-label documents to positive if their cosine similarities are in the top-$\phi$\%. 

\begin{algorithm}[t]
\SetAlgoLined
\KwIn{Unlabeled documents $\{\x^\mathrm{U}_i\}_{i=1}^{n_\mathrm{U}}$, keyword set $ W:=\{w_j\}_{j=1}^{n_\mathrm{W}}$, threshold $\phi\in[0,100]$, weight of original keywords $\alpha\in \mathbb{N}$, sampling factor $\gamma \in \mathbb{N}$.}
\KwOut{two sets of pseudo-labeled documents $X_\mathrm{CP}$ and $X_\mathrm{CN}$}
\BlankLine
$X_\mathrm{CP} := \{ \}$;
$X_\mathrm{CN} := \{ \}$ 
\BlankLine
\ForEach{$w_j$}
{Add additional ($\alpha$-1) of $w_j$ to $W$\;
Find top-$\gamma$ similar words to $w_j$ $W_{sim}:=\{w_j^i\}_{i=1}^{\gamma}$\;
Add $W_{sim}$ to $W$\;}
Compute similarity $s_i$ between one document that merges all keywords $W=\{w_j\}_{j=1}^{(\alpha+\gamma)n_\mathrm{W}}$ and each document $\x^\mathrm{U}_i$\; 
\ForEach{$\x^\mathrm{U}_i$}{
\uIf{$s_i > 0$ and in the top-$\phi$(\%)}
{Add $\x^\mathrm{U}_i$ to $X_\mathrm{CP}$\;}
\Else{Add $\x^\mathrm{U}_i$ to $X_\mathrm{CN}$\;}
}
\Return $X_\mathrm{CP}$, $X_\mathrm{CN}$
 \caption{A pseudo-labeling algorithm}
 \label{alg:pseudo}
\end{algorithm}

\subsection{AUC Optimization from Corrupted Labels}\label{sect:AUC-sym}
After applying Algorithm~\ref{alg:pseudo}, we can obtain two sets of data, which we call them a corrupted positive set and a corrupted negative set. 
Let us define $\theta, \, \theta' \in [0,1]$, which indicate the proportion of positive data for the corrupted positive and negative sets, respectively.
We assume a mild condition that our pseudo-labeling method can split the data such that the corrupted positive set has a higher proportion of positive data than another set, i.e., $\theta > \theta'$. Then, we formulate the data from the two sets as
\begin{align*}
 X_\mathrm{CP}&:= \{\x^\mathrm{CP}_i\}_{i=1}^{n_\mathrm{CP}} \stackrel{\mathrm{i.i.d.}}{\sim} p_\theta(\x) \text{,}\\
 X_{\mathrm{CN}}&:=  \{\x^\mathrm{CN}_j\}_{j=1}^{n_\mathrm{CN}}\stackrel{\mathrm{i.i.d.}}{\sim} p_{\theta'}(\x) \text{,}
\end{align*}
where
\begin{align*}
p_\theta(\x)&:= \theta p(\x|y=1)+(1-\theta) p(\x|y=-1) \text{,}\\
 p_{\theta'}(\x)&:=  \theta' p(\x|y=1)+(1-\theta') p(\x|y=-1) \text{.}
\end{align*}
 $X_\mathrm{CP}$ and $X_\mathrm{CN}$ are the sets of corrupted positive and corrupted negative data, respectively. We utilize the fact that optimizing AUC can be performed effectively even when the data is highly corrupted by using a symmetric loss function~\citep{charoenphakdee2019symmetric}. More specifically, let us consider a corrupted AUC risk. First, let us define $ \mathbb{E}_{\theta}$ and $ \mathbb{E}_{\theta'}$ as the expectation over $p_{\theta}$ and $p_{\theta'}$, respectively. A corrupted AUC risk is defined as 
\begin{align}\label{eq:corrupted-AUC}
R^\ell_{\mathrm{AUC}\text{-}\mathrm{Corr}}(g) = \mathbb{E}_{\theta}[\mathbb{E}_{\theta'}[\ell(g(\x)-g(\x')))]]\text{.}
\end{align}
Intuitively, the risk \eqref{eq:corrupted-AUC} is minimized by a function $g$ that gives a higher value on corrupted positive data over corrupted negative data. The following theorem establishes the relationship between the clean AUC risk and corrupted AUC risk.

\begin{theorem}[\citet{charoenphakdee2019symmetric}]\label{AUC-general}
Let $\varphi^\ell(\x,\x') = \ell(g(\x)-g(\x')) + \ell(g(\x')-g(\x)) $. Then $R^\ell_{\mathrm{AUC}\text{-}\mathrm{Corr}}(g)$ can be expressed as
\begin{align*}
\begin{split}
&R^\ell_{\mathrm{AUC}\text{-}\mathrm{Corr}}(g)  = (\theta-\theta')R^\ell_{\mathrm{AUC}}(g) \\ &  + \underbrace{(1-\theta)\theta' \Epos[\Eneg[\varphi^\ell(\x^\mathrm{P},\x^\mathrm{N})]]}_\textup{Excessive term} \\ &+ \underbrace{\frac{\theta\theta'}{2} \E_{\mathrm{P'}}[\Epos[\varphi^\ell(\x^{\mathrm{P'}},\x^\mathrm{P})]]}_\textup{Excessive term} \\ & + \underbrace{\frac{(1-\theta)(1-\theta')}{2}   \E_{\mathrm{N'}}[\Eneg[\varphi^\ell(\x^{\mathrm{N'}},\x^\mathrm{N})]]}_\textup{Excessive term}\text{.}
\end{split}
\end{align*}
\end{theorem}

Theorem~\ref{AUC-general} shows that minimizing the corrupted risk $R^\ell_{\mathrm{AUC}\text{-}\mathrm{Corr}}(g)$ implies minimizing both the clean risk $R^\ell_{\mathrm{AUC}}(g)$ and excessive terms. However, since we only aim to minimize the clean risk, minimizing the corrupted risk may not effectively minimize the clean risk since excessive terms can be minimized instead and lead to overfitting. ~\citet{charoenphakdee2019symmetric} showed that with a specific class of losses, excessive terms become constant. The following theorem suggests that by using a family of symmetric losses, the corrupted AUC risk can be expressed as an affine transformation of the clean AUC risk. 
\begin{theorem}[\citet{charoenphakdee2019symmetric}]\label{AUC-symmetric}
Let $\sell$ be a symmetric loss such that $\sell(z)+\sell(-z) = K$, where $K$ is a positive constant. Then, $R^{\sell}_{\mathrm{AUC}\text{-}\mathrm{Corr}}(g)$ can be expressed as
\begin{align*}
R^{\sell}_{\mathrm{AUC}\text{-}\mathrm{Corr}}(g)  &= (\theta-\theta')R^{\sell}_{\mathrm{AUC}}(g) \\  &\quad+  \frac{K (1  -\theta + \theta')}{2} \text{.}
\end{align*}
\end{theorem}
According to Theorem~\ref{AUC-symmetric}, we can optimize AUC by using a symmetric loss \emph{without estimating $\theta$ and $\theta'$} and do not suffer from excessive terms. This insight was also suggested by~\citet{menon2015}, but they focused on more general losses, while it has been shown that using a symmetric loss can be preferable~\citep{charoenphakdee2019symmetric}. Our experiments also support that using a symmetric loss is preferable, while using a non-symmetric loss still provided reasonable performance. 
 Similarly to AUC, we can also maximize the balanced accuracy effectively with a symmetric loss (see~\citet{van2015average,charoenphakdee2019symmetric} for more details on the balanced accuracy). Since the result of~\citet{charoenphakdee2019symmetric} only suggests that the minimizer of the corrupted risk and clean risk are identical, we provide a theory to explain why a more accurate pseudo-labeling algorithm can improve the performance by proving an estimation error bound in Section~\ref{sec:bound}.

\subsection{Optimizing Accuracy and $\mathrm{F}_{1}$-measure}\label{sec:eva-metric}
Here, we review other evaluation metrics and discuss how to adjust our framework for optimizing other evaluation metrics. 
Many evaluation metrics can be optimized if a suitable threshold and $p(y=1|\x)$ are known~\citep{yan2018binary}, e.g., $\mathrm{sign}[p(y=1|x)-\frac{1}{2}]$ is the Bayes-optimal solution for the accuracy.
It is known that the Bayes-optimal solution of AUC maximization is any function that has a strictly monotonic relationship with $p(y=1|\x)$~\citep{menon2016bipartite}.
Therefore, finding an appropriate threshold with an AUC maximizer can give an effective classifier~\citep{narasimhan2013relationship}.

For the accuracy and $\mathrm{F}_{1}$-measure, we propose to optimize AUC first and then adjust the threshold to optimize other metrics.
It is important to emphasize that a big challenge behind this problem is that \emph{there is no information about the threshold from given data since all given data are unlabeled.}

Without additional assumptions, we argue that it is \emph{theoretically impossible} to draw an optimal threshold to optimize the accuracy and $\mathrm{F}_{1}$-measure for this problem.
Unlike supervised-learning, where we have positive and negative data, the threshold information is provided from labels and we can draw the decision boundary accordingly. On the other hand, one can learn a bipartite ranking function effectively by AUC optimization in our setting, which suggests the idea to learn a reliable ranking function first, and then adjust the threshold.

Suppose that we know the proportion of positive data $\pi$ in the training data and the class prior of the training and test data are identical. 
A reasonable threshold~$\beta \in \R$ can be given as
\begin{align}\label{eq:thr}
    \pi = \int \mathrm{sign}(g(\x)-\beta) p_{\pi}(\x) d\x.
\end{align}

Intuitively, we use a threshold $\beta$ that decides the top proportion $\pi$ as positives and negative otherwise, which can optimize the accuracy effectively.
This threshold is known as a precision-recall breakeven point, where it is the point where the precision equals to recall (see~\citet{kato2018learning} for its proof). Therefore, it is a reasonable threshold for the $\mathrm{F}_{1}$-measure since it is a harmonic mean of precision and recall. 
With unlabeled documents and $\pi$, we can decide $\beta$ that satisfies the empirical version of Eq.~\eqref{eq:thr}. 

Existing methods, which may learn a threshold without knowing $\pi$, may actually learn a wrong threshold. 
Table~\ref{tab:failure} illustrates the failure of a default threshold when learning a classifier in noisy environments and Table~\ref{tab:rslt} shows that much better performance on $\mathrm{F}_{1}$-measure and accuracy can be recover by adjusting the threshold with Eq.~\eqref{eq:thr}.
Regarding the assumption that training and test class priors are identical, it is the common assumption if we want to use these metrics for evaluation and is implicitly assumed in the existing work~\citep{jin2017combining}.
If the class priors are not identical, we suggest not to use these metrics since it can be misleading because a good $\mathrm{F}_{1}$-measure in training data cannot guarantee a good $\mathrm{F}_{1}$-measure in the test data.
In real-world applications, such situations may occur when we do not collect unlabeled data directly from the test environment or the test environment is prone to be changed over time. 
If it is impossible to estimate $\pi$ or the test condition can be changed, we suggest to use other evaluation metrics such as AUC or precision at $k$.

\section{Theoretical Analysis}\label{sec:bound}
In this section, we provide a theoretical guarantee for our method.
Although it has been shown that the risk minimizer of corrupted data is identical to that of clean data~\citep{charoenphakdee2019symmetric}, the existing work cannot explain why a good pseudo-labeling algorithm can give a better performance.
Throughout this section, we assume that an output of a loss function $\sell$ is bounded by $[0,K]$ without loss of generality for a nonnegative symmetric loss. 
Here, we prove an estimation error bound, which suggests that a larger gap between $\theta-\theta'$ leads to a tighter bound and thus faster convergence. 
Note that the gap is largest when we have clean positive data (i.e., $\theta=1$) and clean negative data (i.e., $\theta'=0$). 
Therefore, a good pseudo-labeling is an algorithm that can divide the data with a large gap (i.e., $\theta-\theta'$ is large). All proofs can be found in Appendix~\ref{apx:est-bound}. 

Let $\hat{g} \in \mathcal{G}$ be a minimizer of the empirical corrupted AUC risk $\widehat{R}^{\sell}_{\mathrm{AUC}\text{-}\mathrm{Corr}}$ in the hypothesis class $\mathcal{G}$. Let $g^* \in \mathcal{G}$ be a minimizer of the expected risk of clean AUC risk $R^{\sell}_{\mathrm{AUC}}$. 
Then, the following lemma establishes the relationship between an estimation error bound of the AUC risk and the uniform deviation bound of the corrupted AUC risk.
\begin{lemma}\label{lem:relation} An estimation error bound of the clean AUC risk $R^{\ell}_{\mathrm{AUC}}$ can be given as follows.
\begin{align*}
     &R^{\sell}_{\mathrm{AUC}}(\hat{g})-R^{\sell}_{\mathrm{AUC}}(g^*) \leq \\ &\frac{2}{\theta-\theta'}\sup\limits_{g \in \mathcal{G}} |R^{\sell}_{\mathrm{AUC}\text{-}\mathrm{Corr}}(g) - \widehat{R}^{\sell}_{\mathrm{AUC}\text{-}\mathrm{Corr}}(g) |.
\end{align*}
\end{lemma}

Next, let $\mathcal{Q}_\mathcal{G}^{\sell}$ be a class of functions mapping $\X^2$ to $[0,K]$ such that $\mathcal{Q}_\mathcal{G}^{\sell}= \{Q: (\x, \x') \to \sell(g(\x)-g(\x')), g \in \mathcal{G}\}$.
Given samples $(x_1,\ldots,x_n)~\in~\mathcal{X}^n$ and  $({x'}_1,\ldots,{x'}_m)~\in~\mathcal{X}^m$ independently and identically drawn from a distribution with densities $\mu$ and $\mu'$, respectively. The empirical bipartite Rademacher complexity~\citep{usunier2005data} of a function class $\mathcal{Q}_\mathcal{G}^{\sell}$ is defined as
\begin{align*}
 &\hat{\mathfrak{R}}_{n, m}(\mathcal{Q}_\mathcal{G}^{\sell}) 
 = \\ &2\mathbb{E}_{\sigma,\upsilon}\sup\limits_{Q \in \mathcal{Q}_\mathcal{G}^{\sell}}\frac{1}{nm}\sum_{i=1}^{n}\sum_{j=1}^{m}\frac{\sigma_i+\upsilon_j}{2} \left[ Q(\x_i, {\x'}_j)\right],
%\ell(g(x_i)-g({x'}_j))
\end{align*}
where the inner expectation is taken over $\sigma=(\sigma_1, \ldots , \sigma_n)$ and $\upsilon=(\upsilon_1, \ldots , \upsilon_m)$ which are independent random variables taking values in $\{-1, +1\}$ uniformly.

% Our analysis employs the Rademacher complexity, which is defined as follows.
% \begin{definition}[Rademacher complexity \cite{bartlett2002rademacher}]\label{rademacher}
% \rm{Let $\mathcal{G}$ be a set of real-valued hypotheses defined over a set $\mathcal{X}$. Given a sample ($x_1,\ldots,x_m$)$ \in \mathcal{X}^m$ drawn i.i.d. from a distribution with density $\mu$, the Rademacher complexity of $\mathcal{G}$ is defined as
% \begin{equation*}
%  \mathfrak{R}_{m}(\mathcal{G}) = \mathbb{E}_{x_1,\ldots,x_m}\mathbb{E}_{\sigma}\left[\sup\limits_{g\in\mathcal{G}}\left(\frac{1}{m}\sum_{j=1}^{m}\sigma_i g(x_i)\right)\right],
% \end{equation*}
% where the inner expectation is taken over $\sigma=(\sigma_1, \ldots , \sigma_m)$ which are independent random variables taking values in $\{-1, +1\}$.} 
% \end{definition}
% Intuitively, the Rademacher complexity measures how much our hypothesis class $\mathcal{G}$ can fit a random noise (see \citet{bartlett2002rademacher} for more details).
% Therefore, the higher the Rademacher complexity is, the more flexible the hypothesis class is (see \citet{bartlett2002rademacher, shalev2014understanding, mohrifoundation} for more details on the measures of the complexity of the hypothesis class).
Then, the bipartite Rademacher complexity can be defined as
\begin{align*}
 \mathfrak{R}_{n, m}(\mathcal{Q}_\mathcal{G}^{\sell}) = \mathbb{E}_{x_1,\ldots,x_n, {x'}_1,\ldots,{x'}_m}\hat{\mathfrak{R}}_{n, m}(\mathcal{Q}_\mathcal{G}^{\sell}).
\end{align*}
Let $\mathfrak{R}^{\mathrm{Corr}}_{n_\mathrm{CP}, n_\mathrm{CN}}$ be the bipartite Rademacher complexities of pseudo-labeled data where $\mu=p_\theta$ and $\mu'=p_\theta'$.
Next, we provide a uniform deviation bound of the corrupted AUC risk. 
%Let $\mathfrak{R}_{\mathrm{CP}}(\mathcal{G})$ and $\mathfrak{R}_{\mathrm{CN}}(\mathcal{G})$ are the Rademacher complexities defined on a corrupted positive sample and  corrupted negative sample, respectively. 
%Then, we obtain the following lemma.
\begin{lemma}[Uniform deviation bound]\label{lem:est-bound} 
%Let $\mathcal{G}$ be a class of functions mapping $\X$ to $\R$. 
For all $Q \in \mathcal{Q}_\mathcal{G}^{\sell}$ and $\delta \in (0,1)$, with probability at least $1-\delta$, we have
\begin{align*}
%   &\sup\limits_{g \in \mathcal{G}} |R^{\sell}_{\mathrm{AUC}\text{-}\mathrm{Corr}}(g) - \widehat{R}^{\sell}_{\mathrm{AUC}\text{-}\mathrm{Corr}}(g) | \leq \\ &K \sqrt{\frac{(n_\mathrm{CP}+n_\mathrm{CN})\log\frac{1}{\delta}}{2n_\mathrm{CP}n_\mathrm{CN}}}
%     \\ &+ 2 L_{\ell} \big[ \mathfrak{R}_{\mathrm{CP}}(\mathcal{G}) +  \mathfrak{R}_{\mathrm{CN}}(\mathcal{G}) \big],
     &\sup\limits_{g \in \mathcal{G}} \left| \corrrisk{g} - \corremprisk{g} \right| \leq    \\ &K \sqrt{\frac{(n_\mathrm{CP}+n_\mathrm{CN})\log\frac{1}{\delta}}{2n_\mathrm{CP}n_\mathrm{CN}}} + 2\mathfrak{R}^{\mathrm{Corr}}_{n_\mathrm{CP}, n_\mathrm{CN}}(\mathcal{Q}_\mathcal{G}^{\sell}).
\end{align*}
where the probability is over repeated sampling of data for evaluating $\widehat{R}^\ell_{\mathrm{AUC}\text{-}\mathrm{Corr}}(g)$.
\end{lemma}
Then, by combining Lemma~\ref{lem:relation} and Lemma~\ref{lem:est-bound}, we obtain the following theorem. 
\begin{theorem}[Estimation error bound]\label{est-bound} %$\mathcal{G}$ be a class of functions. 
For all $Q \in \mathcal{Q}_\mathcal{G}^{\sell}$ and $\delta \in$ (0,1), with probability at least $1-\delta$, we have
\begin{align*}
%   &R^{\sell}_{\mathrm{AUC}}(\hat{g})-R^{\sell}_{\mathrm{AUC}}(g^*) \leq  \\ &\frac{1}{\theta-\theta'} \bigg[ K \sqrt{\frac{2(n_\mathrm{CP}+n_\mathrm{CN})\log\frac{1}{\delta}}{n_\mathrm{CP}n_\mathrm{CN}}}  \bigg]
% \\     &+ \frac{1}{\theta-\theta'}\bigg[4 L_{\ell} \big[ \mathfrak{R}_{\mathrm{CP}}(\mathcal{G}) +  \mathfrak{R}_{\mathrm{CN}}(\mathcal{G}) \big]
   &R^{\sell}_{\mathrm{AUC}}(\hat{g})-R^{\sell}_{\mathrm{AUC}}(g^*) \leq  \\ &\frac{1}{\theta-\theta'} \bigg[ K \sqrt{\frac{2(n_\mathrm{CP}+n_\mathrm{CN})\log\frac{1}{\delta}}{n_\mathrm{CP}n_\mathrm{CN}}}  \bigg]
\\     &+ \frac{4\mathfrak{R}^{\mathrm{Corr}}_{n_\mathrm{CP}, n_\mathrm{CN}}(\mathcal{Q}_\mathcal{G}^{\sell})}{\theta-\theta'},
\end{align*}
where the probability is over repeated sampling of $X_\mathrm{CP}$ and $X_\mathrm{CN}$ for training $\hat{g}$.
\end{theorem}
Theorem~\ref{est-bound} shows that a gap $\frac{1}{\theta-\theta'}$ affects the tightness of the bound. When $\theta$ and $\theta'$ are close to each other, i.e., when the data is highly corrupted, the bound becomes loose. This illustrates the difficulty of the task when pseudo-labeling algorithm performs poorly and we may need more data. 
Nevertheless, as long as $\theta > \theta'$, with all parametric models with their norm is bounded such as neural networks with weight decay or kernel model, our learning framework is consistent, i.e., the estimation error converges to zero as $n_\mathrm{CP},n_\mathrm{CN} \to \infty$. 

% \begin{proof}
%  \textcolor{red}{must be finished on ASAP}
% \end{proof}
% \textcolor{red}{talk about optimal convergence rate and cite mendelson paper}

\section{Experimental Results}
We present experimental results in this section with evaluation metrics include AUC, macro $\mathrm{F}_{1}$-measure, accuracy (ACC) and precision at $100$ (Prec@$100$). Prec@$100$ is the ratio of the true positive data over in the top-$100$ ranking score.

\begin{table}[t]
    \centering
    \caption{Keywords for each dataset}\label{tab:keyword}
    \resizebox{\columnwidth}{!}{\begin{tabular}{c|c}
    \hline
        Dataset & Keywords \\ \hline 
        \multirow{2}{*}{$\mathsf{Subj}$} & wonderful terrible feel happy ugly even horrible   \\
        &  interesting funny dramatic romantic compassionate  \\ \hline
        \multirow{2}{*}{$\mathsf{Custrev}$} & easy excellent nice great good love amazing best   \\
        &  awesome perfect
            definitely better happy compassionate  \\ \hline
       $\mathsf{MPQA}$ & support hope help    good great love  \\ \hline
       $\mathsf{AYI}$ & great best excellent friendly awesome nice amazing  \\ \hline
       $\mathsf{20NG}$ & sports baseball hockey  \\ \hline
    \end{tabular}}
    \label{tab:pseudo}
    % \vspace{-0.8in}
\end{table}

\subsection{Experiment Setup}
\textbf{Datasets:} We used five datasets, namely the Subjectivity dataset~\cite{Pang+Lee:04a} ($\mathsf{Subj}$), Customer reviews~\citep{hu2004mining} ($\mathsf{Custrev}$), Opinion mining in MPQA corpus ($\mathsf{MPQA}$), Product reviews from Amazon, Yelp, and IMDb ($\mathsf{AYI}$), 20 Newsgroups~\cite{Lang95} ($\mathsf{20NG}$) with \emph{baseball} and \emph{hockey} groups as positive. More information on the datasets can be found in Appendix~\ref{apx:exp-info}. Keywords for each dataset are shown in Table~\ref{tab:keyword}.
\begin{table}
\centering
\caption{Failure of optimizing $\mathrm{F}_{1}$-measure if thresholds are not adjusted. Full results can be found in Appendix~\ref{apx:add-exp}. By adjusting the threshold in Table~\ref{tab:rslt} can substantially improve the performance.}\label{tab:failure}
\resizebox{\columnwidth}{!}{\begin{tabular} { |c|c|c|c|c|}
\hline
Methods& $\mathsf{Subj}$& $\mathsf{MPQA}$& $\mathsf{AYI}$& $\mathsf{20NG}$\\ \hline
Maxent &63.4 (0.31) &50.1 (0.22) &42.5 (0.35) &47.4 (0.05) \\ 
NB &73.7 (0.23) &53.8 (0.22) &65.8 (0.42) &23.7 (0.25)  \\ 
RandomForest &33.3 (0.00) & 43.5 (0.20) &35.0 (0.20) &47.2 (0.00) \\ 
KNN &43.6 (0.23) & 51.0 (0.16) &61.6 (0.43) &84.3 (0.26) \\ 
\hline 
\end{tabular}}
\end{table}
\noindent \textbf{Common setup:} First, we need to pseudo-label documents. We used Algorithm~\ref{alg:pseudo} to feed pseudo-labeled documents to all methods, where $\phi$ is set to $90$. ($\alpha$, $\gamma$) were set to ($3$,$5$) for all datasets except $\mathsf{Subj}$, which was ($1$,$50$). We used $50$-dimensional features for GloVe word embeddings. For ACC and $\mathrm{F}_{1}$, we gave true thresholds $\pi$ to all methods to see the top performance of each method. We also provide the results with varying thresholds and heuristic thresholds in Appendix~\ref{apx:add-exp}, where the trends of performance for each method do not differ much from Table~\ref{tab:rslt}. Note that AUC and Prec@$100$ do not use a threshold to evaluate. The experimental results are reported in the mean value and standard error of 20 trials. "N/A" indicates that an algorithm is not terminated due to too many number of vocabularies.

\noindent \textbf{Baselines:} We compared our method with three categories of baselines: the text-feature, GloVe-feature, and zero-shot baselines. 
For the text-feature baselines, we use naive Bayes (NB), the maximum entropy model (Maxent), and naive Bayes that is implemented for learning from positive and unlabeled data (PU-NB). All implementations were from Natural Language Toolkit (NLTK)~\citep{loper2002nltk}. For the GloVe-feature baselines, we used mean word vectors as features and employed a random forest (RandomForest) and K-nearest neighbors (KNN), which were implemented by Scikit-learn~\citep{pedregosa2011scikit}.
Finally, the zero-shot baselines did not use unlabeled data but simply GloVe to rank the score of a document (GloVeRanking) and keyword voting (Voting). We also showed the performance when fully-labeled data are given as references. O-Maxent is a maxent model with fully-labeled data and O-Sigmoid is our framework that skips the pseudo-labeling step and uses fully-labeled data for AUC optimization.
\begin{table*}
\centering
\caption{Mean value and standard error of 20 trials for the AUC, $\mathrm{F}_{1}$-measure, accuracy (ACC), and precision at $100$ (Prec@$100$) of learning from relavant keywords and unlabeled documents. Outperforming methods are highlighted in boldface using one-sided t-test with the significance level of 5\%. The sigmoid loss is symmetric while the logistic loss is non-symmetric. }\label{tab:rslt}
\resizebox{\textwidth}{!}{\begin{tabular} {|c|c|c|c|c|c|c|c|c|c|c||c|c|}
\hline
\multirow{2}{*}{Dataset}&\multirow{2}{*}{Evaluation}&\multicolumn{2}{c|}{Proposed framework} & \multicolumn{3}{c|}{Text-feature baselines} & \multicolumn{2}{c|}{GloVe-feature baselines} & \multicolumn{2}{c||}{Zero-shot baselines}& \multicolumn{2}{c|}{Oracle} \\
&& Sigmoid& Logistic& PU-NB& NB& Maxent& RandomForest& KNN & GloVeRanking& Voting& O-Maxent& O-Sigmoid\\ \hline
\multirow{4}{*}{$\mathsf{Subj}$}&AUC & \textbf{88.1 (0.35)} &84.1 (0.30) &55.4 (0.13) &85.0 (0.18) &84.6 (0.20) &82.4 (0.27) &73.6 (0.29) &81.7 (0.19) &70.2 (0.24) &97.4 (0.06) &93.6 (0.11)\\ 
&$\mathrm{F}_{1}$ & \textbf{80.1 (0.38)} &76.0 (0.32) &47.1 (0.20) &76.3 (0.16) &76.3 (0.24) &75.1 (0.27) &63.6 (0.32) &74.5 (0.13) &63.5 (0.18) &92.0 (0.13) &86.4 (0.14)\\ 
&ACC & \textbf{80.1 (0.38)} &76.0 (0.32) &55.0 (0.13) &76.3 (0.16) &76.3 (0.24) &75.1 (0.27) &65.0 (0.28) &74.5 (0.13) &64.1 (0.18) &92.0 (0.13) &86.4 (0.14)\\ 
&Prec@$100$ & \textbf{96.3 (0.60)} & \textbf{95.1 (0.60)} &0.9 (0.09) & \textbf{95.9 (0.33)} &94.7 (0.39) &93.2 (0.50) &91.5 (0.59) & \textbf{95.2 (0.54)} &85.8 (0.93) &99.3 (0.15) &97.8 (0.27)\\ 
\hline 
\multirow{4}{*}{$\mathsf{Custrev}$}&AUC &71.2 (0.34) &70.7 (0.34) &59.7 (0.32) & \textbf{74.2 (0.35)} &69.9 (0.46) &62.6 (0.47) &67.9 (0.40) &55.3 (0.39) &67.4 (0.32) &75.0(0.38) & 78.5(0.30)\\ 
&$\mathrm{F}_{1}$ & \textbf{63.6 (0.41)} & \textbf{63.1 (0.31)} &58.4 (0.35) & \textbf{63.3 (0.49)} &62.1 (0.43) &58.3 (0.44) &60.8 (0.33) &53.0 (0.40) &38.9 (0.00) &68.5 (0.42) & 70.2 (0.29)\\ 
&ACC &66.5 (0.40) &66.0 (0.31) &64.8 (0.32) & \textbf{69.8 (0.37)} &65.8 (0.39) &61.9 (0.40) &66.6 (0.29) &56.6 (0.37) &63.7 (0.00) &71.4 (0.41)  & 72.5 (0.28)\\ 
&Prec@$100$ &91.2 (0.46) &91.5 (0.49) & \textbf{99.2 (0.18)} &89.8 (0.55) &91.2 (0.57) &80.9 (0.49) &86.3 (0.85) &75.1 (0.81) &87.3 (0.66) & 82.2 (0.35) & 86.9 (0.28)\\ 
\hline 
\multirow{4}{*}{$\mathsf{MPQA}$}&AUC & \textbf{80.4 (0.44)} &78.7 (0.37) &52.1 (0.27) &56.4 (0.31) &56.7 (0.23) &69.1 (0.55) &60.1 (0.23) &63.6 (0.26) &56.0 (0.12) &78.3 (0.25) &86.8 (0.18)\\ 
&$\mathrm{F}_{1}$ & \textbf{71.7 (0.44)} &69.8 (0.31) &46.7 (0.23) &54.3 (0.28) &53.1 (0.20) &62.4 (0.45) &23.8 (0.00) &57.5 (0.17) &23.8 (0.00) &69.8 (0.19) &77.9 (0.22)\\ 
&ACC & \textbf{75.6 (0.39)} &74.0 (0.27) &47.1 (0.24) &62.4 (0.28) &58.4 (0.17) &67.4 (0.39) &31.2 (0.00) &63.3 (0.17) &31.2 (0.00) &72.8 (0.20) &81.0 (0.19)\\ 
&Prec@$100$ & \textbf{81.5 (0.97)} &77.5 (1.02) &10.8 (3.37) &69.5 (0.86) &63.8 (1.93) &76.9 (1.06) &78.7 (0.80) &50.6 (0.60) &74.7 (0.69) &94.8 (0.46) &90.5 (0.52)\\ 
\hline 
\multirow{4}{*}{$\mathsf{AYI}$}&AUC & \textbf{76.0 (0.41)} & \textbf{75.6 (0.43)} &60.5 (0.39) &71.2 (0.41) &60.7 (0.46) &70.1 (0.55) &72.5 (0.39) &62.4 (0.53) &61.0 (0.33) &84.6 (0.32) &81.1 (0.40)\\ 
&$\mathrm{F}_{1}$ & \textbf{69.3 (0.36)} & \textbf{68.8 (0.40)} &58.9 (0.47) &61.6 (0.38) &56.6 (0.36) &64.5 (0.53) &65.5 (0.52) &58.7 (0.51) &33.5 (0.00) &76.8 (0.37) &73.0 (0.39)\\ 
&ACC & \textbf{69.3 (0.36)} & \textbf{68.8 (0.40)} &60.1 (0.41) &62.5 (0.35) &56.8 (0.36) &64.6 (0.53) &65.8 (0.44) &58.7 (0.51) &50.5 (0.00) &76.9 (0.37) &73.0 (0.39)\\ 
&Prec@$100$ & \textbf{87.5 (0.55)} & \textbf{87.5 (0.62)} &74.5 (2.20) &85.1 (0.71) &70.2 (1.00) &77.2 (0.99) &82.5 (0.69) &72.4 (0.91) &79.2 (0.87) &95.6 (0.47) &90.1 (0.73)\\ 
\hline 
\multirow{4}{*}{$\mathsf{20NG}$}&AUC &96.4 (0.12) &96.0 (0.15) & N/A &77.1 (0.21) &57.6 (0.32) & \textbf{96.8 (0.16)} &94.7 (0.16) &95.0 (0.17) &62.9 (0.22) &65.5 (0.46) &99.0 (0.05)\\ 
&$\mathrm{F}_{1}$ & \textbf{90.8 (0.20)} & \textbf{90.6 (0.21)} & N/A &58.4 (0.22) &52.4 (0.25) &89.6 (0.28) &86.7 (0.59) & \textbf{90.5 (0.18)} &9.6 (0.00) &56.8 (0.29) &94.1 (0.14)\\ 
&ACC & \textbf{96.5 (0.08)} & \textbf{96.4 (0.08)} & N/A &70.2 (0.31) &81.6 (0.11) &96.1 (0.10) &94.5 (0.35) & \textbf{96.4 (0.07)} &10.6 (0.00) &83.5 (0.11) &97.8 (0.05)\\ 
&Prec@$100$ & \textbf{99.5 (0.15)} & \textbf{99.1 (0.24)} & N/A &0.4 (0.11) &17.6 (0.77) & \textbf{99.5 (0.15)} &97.6 (0.38) &97.5 (0.28) &85.2 (1.03) &32.0 (1.31) &99.9 (0.07)\\ 
\hline 
\end{tabular}}
\end{table*}
\begin{table*}
\centering
\caption{Mean $\mathrm{F}_{1}$-measure and standard error of 20 trials with varying thresholds with different $\hat{\pi}$ in Eq.~\eqref{eq:thr}.}
\label{tab:thrsld}
\resizebox{\textwidth}{!}{\begin{tabular} { |c|c|c|c|c|c|c|c|c|c|c|c|}
\hline
Dataset & Methods& 0.05& 0.1& 0.2& 0.3& 0.4& 0.5& 0.6& 0.7& 0.8& 0.9\\ \hline
\multirow{4}{*}{$\mathsf{Subj}$}&Sigmoid &43.3 (0.17) &51.2 (0.24) &63.9 (0.34) &72.5 (0.27) &77.9 (0.37) & \textbf{80.1 (0.38)} &78.7 (0.32) &73.7 (0.29) &64.7 (0.32) &51.8 (0.30)\\ 
&Maxent &37.1 (0.15) &41.9 (0.18) &52.5 (0.27) &62.1 (0.19) &70.6 (0.27) & \textbf{76.3 (0.24)} &74.5 (0.25) &64.1 (0.25) &50.3 (0.22) &39.3 (0.13)\\ 
&RandomForest &42.1 (0.22) &49.9 (0.15) &61.6 (0.22) &69.1 (0.26) &73.5 (0.29) & \textbf{75.1 (0.27)} &73.5 (0.17) &69.0 (0.21) &61.2 (0.24) &49.6 (0.23)\\ 
&GloVe Ranking &43.1 (0.24) &50.5 (0.25) &61.9 (0.25) &69.3 (0.27) &73.6 (0.18) & \textbf{74.5 (0.13)} &72.8 (0.18) &68.0 (0.16) &60.6 (0.23) &49.4 (0.22)\\ 
\hline 
\multirow{4}{*}{$\mathsf{20NG}$}&Sigmoid &79.9 (0.23) & \textbf{91.2 (0.18)} &78.4 (0.20) &68.2 (0.15) &60.0 (0.14) &52.7 (0.13) &45.3 (0.15) &37.8 (0.13) &29.7 (0.13) &20.5 (0.13)\\ 
&Maxent &51.5 (0.19) & \textbf{52.3 (0.24)} &51.5 (0.21) &49.5 (0.16) &46.6 (0.15) &43.0 (0.16) &38.1 (0.14) &32.9 (0.15) &26.3 (0.16) &19.0 (0.13)\\ 
&RandomForest &79.4 (0.33) & \textbf{89.8 (0.29)} &78.5 (0.17) &68.2 (0.15) &59.8 (0.16) &52.6 (0.13) &45.4 (0.12) &37.7 (0.16) &29.6 (0.13) &20.3 (0.10)\\ 
&GloVe Ranking &79.2 (0.34) & \textbf{90.7 (0.17)} &78.2 (0.22) &67.9 (0.19) &59.9 (0.14) &52.3 (0.11) &45.0 (0.10) &37.6 (0.10) &29.3 (0.11) &20.1 (0.13)\\ 
\hline 
\end{tabular}}
\end{table*}

\noindent \textbf{Proposed methods:} 
For the AUC optimization part, we used the recurrent convolutional neural networks (RCNN) model~\cite{lai2015recurrent} with two layer long short-term memory (LSTM)~\cite{hochreiter1997long}.
We used Adam~\citep{adam} as an optimization method. We used the symmetric sigmoid loss (Sigmoid). We also show the results of the non-symmetric logistic loss (Logistic) to validate the improvement from using a symmetric loss. Implementation details are provided in Appendix~\ref{apx:exp-info}.

\subsection{End-to-end Classification Performance}
Table~\ref{tab:rslt} shows the classification performance. 
It can be observed that our proposed framework outperforms other baselines in many cases.
Moreover, the sigmoid loss (Sigmoid) outperforms the logstic loss (Logistic), which agrees with Theorem~\ref{AUC-symmetric}. Nevertheless, the performance of the logistic loss is desirable compared with baselines. It can be observed that GloveRanking, which does not use unlabeled data can outperform many baselines in $\mathsf{20NG}$. This can be due to the fact that pseudo-labeling data were corrupted and degraded the performance of a classifier heavily.
In Table \ref{tab:rslt}, for $\mathsf{MPQA}$ and $\mathsf{20NG}$, our proposed framework with symmetric losses is able to outperform O-Maxent in the $\mathrm{F}_{1}$-measure, AUC score, and accuracy without having access to labels. One possible reason that O-Maxent does not perform well on $\mathsf{20NG}$ is class imbalance.

Table~\ref{tab:thrsld} shows the accuracy and  $\mathrm{F}_{1}$-measure with varying thresholds.
The true class prior of $\mathsf{Subj}$ and $\mathsf{20NG}$ were $0.5$ and $0.11$, respectively. It can be observed that the threshold based on Eq.~\eqref{eq:thr} gives a desirable performance when $\hat{\pi}$ is close to $\pi$.
%It is highly challenging to estimate the $\pi$ of $\mathsf{20NG}$ without any prior knowledge on datasets. 
For Prec@$100$, we can see that Prec@$100$ for $\mathsf{20NG}$ for the sigmoid loss in our framework and RandomForest is $99.5$, without any labeled data, which indicates that we may pick top-$100$ documents as positive with almost perfect precision.

\section{Conclusion}
We proposed a theoretically-grounded framework for learning from relevant keywords and unlabeled documents. 
Our framework is highly flexible and can guarantee any heuristic pseudo-labeling method as long as an algorithm can divide unlabeled documents into two sets of data with different proportions of positive data. Experiments showed the usefulness of the proposed method.
\section*{Acknowledgement}
NC was supported by MEXT scholarship and JST AIP challenge. MS was supported by JST CREST Grant Number JPMJCR1403. Part of the research was done while NC was visiting Chulalongkorn University with the support of U. Tokyo IST Research Internship Program.

\bibliographystyle{acl_natbib}

\appendix
\onecolumn
\section{Proof of the Estimation Error Bound}\label{apx:est-bound}
In this section, we prove the estimation error bound of AUC optimization from corrupted labels.

\subsection{Proof of Lemma~\ref{lem:relation}}
With the result of~\citet{charoenphakdee2019symmetric}, we can relate the expected clean AUC risk and corrupted risk as
\begin{align}\label{eq:clean-corrupt-relation}
R^{\sell}_{\mathrm{AUC}\text{-}\mathrm{Corr}}(g)  &= (\theta-\theta')R^{\sell}_{\mathrm{AUC}}(g) +  \frac{K (1  -\theta + \theta')}{2} \text{.}
\end{align}
Next, we derive the estimation error bound of the corrupted risk. Note that $\hat{g}$ is an empirical minimizer of the corrupted AUC risk, i.e., $\hat{g}=\argmin\limits_{g \in \mathcal{G}} \corremprisk{g}$ and therefore
\begin{align*}
    \corremprisk{g^*}-\corremprisk{\hat{g}} \geq 0
\end{align*}

Then, we can write the following:
\begin{align*}
    \corrrisk{\hat{g}} &= \corrrisk{\hat{g}} - \corrrisk{g^*} + \corrrisk{g^*} \\
    & \leq \corrrisk{\hat{g}} - \corrrisk{g^*} + \corrrisk{g^*} + \left[ \corremprisk{g^*}-\corremprisk{\hat{g}} \right]\\
    &= \left[ \corrrisk{\hat{g}} -\corremprisk{\hat{g}} \right] + \left[ \corrrisk{g^*} -\corremprisk{g^*} \right] + \corrrisk{g^*} \\
    &\leq 2\sup\limits_{g \in \mathcal{G}} |R^{\sell}_{\mathrm{AUC}\text{-}\mathrm{Corr}}(g) - \widehat{R}^{\sell}_{\mathrm{AUC}\text{-}\mathrm{Corr}}(g) | + \corrrisk{g^*}
\end{align*}

Thus, the estimation error bound of the corrupted AUC risk can be given as
\begin{align}\label{ineq:corr-est}
    \corrrisk{\hat{g}}-\corrrisk{g^*} &\leq 2\sup\limits_{g \in \mathcal{G}} \left|R^{\sell}_{\mathrm{AUC}\text{-}\mathrm{Corr}}(g) - \widehat{R}^{\sell}_{\mathrm{AUC}\text{-}\mathrm{Corr}}(g) \right|
\end{align}
Let $\mathrm{Const} = \frac{K (1  -\theta + \theta')}{2}$, by applying Eq.~\eqref{eq:clean-corrupt-relation} to the left side of Ineq.~\eqref{ineq:corr-est}, we can obtain the following:
\begin{align*}
     \left[ (\theta-\theta')R^{\sell}_{\mathrm{AUC}}(\hat{g}) +  \mathrm{Const}\right] - \left[ (\theta-\theta')R^{\sell}_{\mathrm{AUC}}(g^*) +  \mathrm{Const} \right] &\leq 2\sup\limits_{g \in \mathcal{G}} \left|R^{\sell}_{\mathrm{AUC}\text{-}\mathrm{Corr}}(g) - \widehat{R}^{\sell}_{\mathrm{AUC}\text{-}\mathrm{Corr}}(g) \right|
\end{align*}
Since the constant terms cancel itself and we can multiply both sides by $(\theta-\theta')$ since $\theta > \theta'$, we obtain the following: 
\begin{align}\label{ineq:est-bound-proof}
     R^{\sell}_{\mathrm{AUC}}(\hat{g}) - R^{\sell}_{\mathrm{AUC}}(g^*)  &\leq \frac{2}{\theta-\theta'}\sup\limits_{g \in \mathcal{G}} \left|R^{\sell}_{\mathrm{AUC}\text{-}\mathrm{Corr}}(g) - \widehat{R}^{\sell}_{\mathrm{AUC}\text{-}\mathrm{Corr}}(g) \right|
\end{align}
Thus, we conclude the proof of Lemma~\ref{lem:relation}.
\subsection{Proof of Lemma~\ref{lem:est-bound}}
Here, we prove the uniform deviation bound of the corrupted AUC risk. We employ the technique from~\citet{usunier2005data} to prove the estimation error bound. Following~\citet{usunier2005data}, we consider another notion of Rademacher complexity for bipartite ranking. First, let us restate the empirical bipartite Rademacher complexity as follows here for convenience.
\begin{definition}[Bipartite empirical Rademacher complexity (a slightly modified version of \citet{usunier2005data})]\label{emp-rademacher-bipartite-apx}
\rm{Let $\mathcal{Q}_\mathcal{G}^{\sell}$ be a class of functions mapping $\X^2$ to $[0,K]$ such that $\mathcal{Q}_\mathcal{G}^{\sell}= \{Q_g^{\sell}: (\x, \x') \to \sell(g(\x)-g(\x')), g \in \mathcal{G}\}$.
Given samples $(x_1,\ldots,x_n)~\in~\mathcal{X}^n$ and  $({x'}_1,\ldots,{x'}_m)~\in~\mathcal{X}^m$ independently and identically drawn from a distribution with densities $\mu$ and $\mu'$, respectively. The empirical bipartite Rademacher complexity of a function class $\mathcal{Q}_\mathcal{G}^{\sell}$ is defined as
\begin{align*}
 \hat{\mathfrak{R}}_{n, m}(\mathcal{Q}_\mathcal{G}^{\sell}) 
 = 2\mathbb{E}_{\sigma,\upsilon}\sup\limits_{Q \in \mathcal{Q}_\mathcal{G}^{\sell}}\frac{1}{nm}\sum_{i=1}^{n}\sum_{j=1}^{m}\frac{\sigma_i+\upsilon_j}{2} \left[ Q(\x_i, {\x'}_j)\right],
%\ell(g(x_i)-g({x'}_j))
\end{align*}
where the inner expectation is taken over $\sigma=(\sigma_1, \ldots , \sigma_n)$ and $\upsilon=(\upsilon_1, \ldots , \upsilon_m)$ which are independent random variables taking values in $\{-1, +1\}$ uniformly.} 
\end{definition}
Then, the bipartite Rademacher complexity can be defined as
\begin{definition}[Bipartite Rademacher complexity \cite{usunier2005data}]\label{rademacher-bipartite}
\rm{Let $\hat{\mathfrak{R}}_{n, m}(\mathcal{Q}_\mathcal{G}^{\sell})$ be an empirical bipartite Rademacher complexity. The bipartite Rademacher complexity of $\mathcal{Q}_\mathcal{G}^{\sell}$ is defined as
\begin{equation*}
 \mathfrak{R}_{n, m}(\mathcal{Q}_\mathcal{G}^{\sell}) = \mathbb{E}_{x_1,\ldots,x_n, {x'}_1,\ldots,{x'}_m}\hat{\mathfrak{R}}_{n, m}(\mathcal{Q}_\mathcal{G}^{\sell}).
\end{equation*}
} 
\end{definition}

 Our next step is similar to the result from Appendix A of~\citet{usunier2005data} et al., the only different is that we assume that an output of a loss function $\sell$ is bounded by $[0,K]$ without loss of generality for nonnegative symmetric losses.
 %Since the change of $\corremprisk{g}$ is not deviate for more than $\frac{K}{n_\mathrm{CP}}+\frac{K}{n_\mathrm{CN}}$ if a data point is replaced by a new data point.
 %Then, 
 By using Mcdiarmid's inequality~\citep{mcdiarmid1989method, usunier2005data}, for $\epsilon \in (0,1]$, we have
 \begin{align*}
     &\mathrm{Pr}\left(\sup\limits_{g \in \mathcal{G}} \left| \corrrisk{g} - \corremprisk{g} \right| - \mathbb{E}_{X_\mathrm{CP} X_\mathrm{CN}} \sup\limits_{g \in \mathcal{G}} \left| \corrrisk{g} - \corremprisk{g} \right| > \epsilon \right) \\ &\leq \exp \left( \frac{-2n_\mathrm{CP}n_\mathrm{CN}\epsilon^2}{K^2(n_\mathrm{CP}+n_\mathrm{CN})}  \right)
 \end{align*}
 Then we apply the inversion technique~\citep{bousquet2003introduction}.
 More specifically, we set 
 \begin{align*}
\exp \left( \frac{-2n_\mathrm{CP}n_\mathrm{CN}\epsilon^2}{K^2(n_\mathrm{CP}+n_\mathrm{CN})}  \right) = \delta     ,
 \end{align*}
  and then solve for $\epsilon$.
 Then, for $\delta \in (0,1]$, the following bound holds with probability $1-\delta$.
 \begin{align*}
     \sup\limits_{g \in \mathcal{G}} \left| \corrrisk{g} - \corremprisk{g} \right| \leq   &\, \mathbb{E}_{X_\mathrm{CP} X_\mathrm{CN}} \sup\limits_{g \in \mathcal{G}} \left| \corrrisk{g} - \corremprisk{g} \right|\\ &+K \sqrt{\frac{(n_\mathrm{CP}+n_\mathrm{CN})\log\frac{1}{\delta}}{2n_\mathrm{CP}n_\mathrm{CN}}}.
 \end{align*}
 The next step is based on symmetrization procedure, which the following result can be obtained directly from~\citet{usunier2005data}.
 \begin{align*}
     \mathbb{E}_{X_\mathrm{CP} X_\mathrm{CN}} \sup\limits_{g \in \mathcal{G}} \left| \corrrisk{g} - \corremprisk{g} \right| \leq 2\mathfrak{R}_{n, m}(\mathcal{Q}_\mathcal{G}^{\sell}) .
 \end{align*}

Thus, we obtain the following bound, with probability $1-\delta$ where the probability is over repeated sampling of data for evaluating $\widehat{R}^\ell_{\mathrm{AUC}\text{-}\mathrm{Corr}}(g)$.
\begin{align}\label{result:lem-uniform}
     \sup\limits_{g \in \mathcal{G}} \left| \corrrisk{g} - \corremprisk{g} \right| \leq   &\, 2\mathfrak{R}^{\mathrm{Corr}}_{n_\mathrm{CP}, n_\mathrm{CN}}(\mathcal{Q}_\mathcal{G}^{\sell})+K \sqrt{\frac{(n_\mathrm{CP}+n_\mathrm{CN})\log\frac{1}{\delta}}{2n_\mathrm{CP}n_\mathrm{CN}}}.
\end{align}
This concludes the proof of Lemma~\ref{lem:est-bound}.
\subsection{Proof of Theorem~\ref{est-bound}}
Once Lemmas~\ref{lem:relation} and \ref{lem:est-bound} are obtained, Theorem~\ref{est-bound} can be derived straightforwardly by combining both lemmas by replacing the supremum term in the right hand side of Ineq.~\eqref{ineq:est-bound-proof} in Lemma~\ref{lem:relation}, which is  $\sup\limits_{g \in \mathcal{G}} \left|R^{\sell}_{\mathrm{AUC}\text{-}\mathrm{Corr}}(g) - \widehat{R}^{\sell}_{\mathrm{AUC}\text{-}\mathrm{Corr}}(g) \right|$, with the right hand side of Ineq.~\eqref{result:lem-uniform} from Lemma~\ref{lem:est-bound}. Then the following result is obtained directly with probability $1-\delta$,
% \begin{align*}
%   &R^{\sell}_{\mathrm{AUC}}(\hat{g})-R^{\sell}_{\mathrm{AUC}}(g^*) \leq  \frac{1}{\theta-\theta'} \bigg[ K \sqrt{\frac{2(n_\mathrm{CP}+n_\mathrm{CN})\log\frac{1}{\delta}}{n_\mathrm{CP}n_\mathrm{CN}}}  \bigg]
% + \frac{1}{\theta-\theta'}\bigg[4 L_{\ell} \big[ \mathfrak{R}_{\mathrm{CP}}(\mathcal{G}) +  \mathfrak{R}_{\mathrm{CN}}(\mathcal{G}) \big] \bigg],
% \end{align*}
\begin{align*}
   &R^{\sell}_{\mathrm{AUC}}(\hat{g})-R^{\sell}_{\mathrm{AUC}}(g^*) \leq  \frac{1}{\theta-\theta'} \bigg[ K \sqrt{\frac{2(n_\mathrm{CP}+n_\mathrm{CN})\log\frac{1}{\delta}}{n_\mathrm{CP}n_\mathrm{CN}}}  \bigg]
+ \frac{4\mathfrak{R}^{\mathrm{Corr}}_{n_\mathrm{CP}, n_\mathrm{CN}}(\mathcal{Q}_\mathcal{G}^{\sell})}{\theta-\theta'},
\end{align*}
where the probability is over repeated sampling of $X_\mathrm{CP}$ and $X_\mathrm{CN}$ for training $\hat{g}$.
\section{Dataset and Implementation details}\label{apx:exp-info}
In this section, we explain more details in experiment.
We split data into 80\% of train data and 20\% of test data respectively.
Since we train each method for each dataset 20 times, we shuffle dataset to make different train data from other trials while every method are trained with same data for each trial.
We used a larger learning rate for small dataset $\mathsf{Custrev}$ and $\mathsf{AYI}$, since the training takes more epoch to finish due to a lower number of mini-batch. More specifically, we used $10^{-4}$ for $\mathsf{Custrev}$ and $\mathsf{AYI}$ and  $10^{-5}$ for other datasets. Apart from that, we used exactly same architecture and hyperparameters for all experiments. We set weight decay factor to $0.003$. Nevertheless, we found that the weight decay is not sensitively affect the performance for a reasonable weight decay that is not too large. We provide the dataset information and the result of Algorithm \ref{alg:pseudo} for each dataset in Table~\ref{tab:datainfo}. Note that we only pseudo-label the train data, not both test data and $\theta, \theta'$ are unknown to learning algorithm.

\begin{table}[htb]
\centering
\caption{Dataset information and result of Algorithm~\ref{alg:pseudo}}
\resizebox{\textwidth}{!}{\begin{tabular} {|c|C|C|C||C|C|C|C|C|}
\hline
\multirow{2}{*}{Dataset}& \multicolumn{3}{c||}{True values} & \multicolumn{5}{c|}{Algorithm \ref{alg:pseudo}}\\
 & \text{Train data} (n_p, n_n) & \text{Test data} (n_p, n_n) &  \pi & (\alpha, \gamma) &  X_\mathrm{CP} (n_{TP}, n_{FP}) &  X_\mathrm{CN} (n_{FN}, n_{TN}) &(\theta, \theta') & \hat{\pi}_{\mathrm{Hue}}  \\ 
\hline
$\mathsf{Subj}$ & (4000, 4000) & (1000, 1000) & 0.50 & (1, 50) & (2688, 1596) & (1312, 2404) & (0.63, 0.35) & 0.54 \\
$\mathsf{Custrev}$ & (1924, 1092) & (481, 274) & 0.65 & (3, 5) & (1007, 237) & (917, 855) & (0.81, 0.52) & 0.41 \\
$\mathsf{MPQA}$ & (2648, 5833) & (663, 1459) & 0.31 & (3, 5) & (328, 90) & (2320, 5743) & (0.78, 0.29) & 0.05 \\
$\mathsf{AYI}$ & (1108, 1089) & (278, 273) & 0.50 & (3, 5) & (268, 90) & (640, 999) & (0.84, 0.39)&  0.16 \\
$\mathsf{20NG}$ & (1594, 13482) & (399, 3371) & 0.11 & (3, 5) & (669, 232) & (925, 13250) & (0.74, 0.07) & 0.06 \\
\hline 
\end{tabular}}
\label{tab:datainfo}
\end{table}

%All datasets are provided in the software link attached to the submission. We will release the download link after the notification (so as the code for the full implementations). 
For baseline methods, we tried to use the standard parameters for all methods unless there are legitimate reasons. For KNN, we tune the number of neighbors to evaluate as $10$ since we observe that the performance is reasonable since too low neighbors may fail miserably and too high neighbors is computationally infeasible. For Maxent model, we used Generalized Iterative Scaling (GIS) method to optimize since we observed that Improved Iterative Scaling (IIS) can lead to numerical instability and cause the program to have an overflow exception. For PositiveNaiveBayes in $\mathsf{20NG}$ dataset, we used "Intel(R) Xeon(R) CPU E5-2450 0 @ 2.10GHz" to run. Unfortunately, we waited for 40 hours and found that the algorithm did not terminate, yet we had to run for 20 trials in order to conduct a one-sided t-test to validate the significance of the method. 
Thus, we report as "N/A".

\section{Additional Experimental Results}\label{apx:add-exp}
In this section, we present more experimental results with different threshold criteria and also visualize the performance of $\mathrm{F}_{1}$ and ACC with varying thresholds.
\subsection{Performance of a Default Threshold}
Table~\ref{tab:default-thr} shows a performance without adjusting a threshold.
\begin{table}[htb]
\centering
\caption{Mean value and standard error of 20 trials for the $\mathrm{F}_{1}$-measure, and accuracy (ACC) of learning from relavant keywords and unlabeled documents \textbf{with the default threshold of each algorithm}. For our framework we simply used $\beta=0.5$. Outperforming methods are highlighted in boldface using one-sided t-test with the significance level of 5\%.}
\resizebox{\textwidth}{!}{\begin{tabular} {|c|c|c|c|c|c|c|c|c|c|c|}
\hline
\multirow{2}{*}{Dataset}&\multirow{2}{*}{Evaluation}&\multicolumn{2}{c|}{Proposed framework} & \multicolumn{3}{c|}{Text-feature baselines} & \multicolumn{2}{c|}{GloVe-feature baselines} & \multicolumn{2}{c|}{Zero-shot baselines} \\
&& Sigmoid& Logistic& PU-NB& NB& Maxent& RandomForest& KNN& GloveRanking& Voting\\ \hline
\multirow{2}{*}{$\mathsf{Subj}$}&$\mathrm{F}_{1}$ &69.0 (1.75) &63.1 (1.55) &45.9 (0.22) & \textbf{73.7 (0.23)} &63.4 (0.31) &33.3 (0.00) &43.6 (0.23) &39.5 (0.13) &56.0 (0.20)\\ 
&ACC &71.4 (1.33) &66.7 (1.06) &55.7 (0.12) & \textbf{74.6 (0.20)} &66.9 (0.23) &50.00 (0.00) &54.7 (0.13) &52.4 (0.08) &59.9 (0.15)\\ 
\hline 
\multirow{2}{*}{$\mathsf{Custrev}$}&$\mathrm{F}_{1}$ &53.9 (0.63) &47.3 (0.48) &57.7 (0.43) & \textbf{66.8 (0.36)} &40.6 (0.28) &43.4 (0.41) &61.9 (0.38) &41.6 (0.22) &62.0 (0.34)\\ 
&ACC &66.0 (0.23) &64.7 (0.18) &67.0 (0.30) & \textbf{68.3 (0.37)} &44.8 (0.20) &45.8 (0.28) &64.1 (0.39) &63.3 (0.13) &63.3 (0.34)\\ 
\hline 
\multirow{2}{*}{$\mathsf{MPQA}$}&$\mathrm{F}_{1}$ & \textbf{67.2 (1.23)} & \textbf{67.1 (0.69)} &38.5 (0.20) &53.8 (0.22) &50.1 (0.22) &43.5 (0.20) &51.0 (0.16) &49.8 (0.20) &53.3 (0.19)\\ 
&ACC & \textbf{73.8 (0.79)} & \textbf{72.2 (0.68)} &39.4 (0.18) &67.3 (0.15) &68.1 (0.13) &69.5 (0.06) &71.4 (0.07) &49.8 (0.20) &71.4 (0.07)\\ 
\hline 
\multirow{2}{*}{$\mathsf{AYI}$}&$\mathrm{F}_{1}$ & \textbf{69.5 (0.35)} &65.5 (0.49) &54.4 (0.49) &65.8 (0.42) &42.5 (0.35) &35.0 (0.20) &61.6 (0.43) &39.1 (0.30) &56.5 (0.44)\\ 
&ACC & \textbf{69.5 (0.35)} &67.0 (0.41) &58.0 (0.32) &65.9 (0.42) &53.7 (0.20) &50.3 (0.09) &64.1 (0.34) &51.6 (0.18) &60.3 (0.33)\\ 
\hline 
\multirow{2}{*}{$\mathsf{20NG}$}&$\mathrm{F}_{1}$ &59.0 (0.92) &51.9 (1.97) &N/A &23.7 (0.25) &47.4 (0.05) &47.2 (0.00) & \textbf{84.3 (0.26)} &54.2 (0.11) &66.9 (0.28)\\ 
&ACC &68.4 (1.19) &58.9 (2.59) &N/A &23.8 (0.26) &89.2 (0.02) &89.4 (0.00) & \textbf{95.1 (0.06)} &62.3 (0.13) &91.0 (0.06)\\ 
\hline 
\end{tabular}}
\label{tab:default-thr}
\end{table}
\pagebreak
\subsection{Heuristic Threshold $\hat{\pi}_{\mathrm{Hue}}$ Based on Algorithm~\ref{alg:pseudo}}
Table~\ref{tab:heu-thr} shows the performance by picking the threshold that is the ratio between pseudo-positive data over unlabeled data.
\begin{table}[htb]
\centering
\caption{Mean value and standard error of 20 trials for the $\mathrm{F}_{1}$-measure, and accuracy (ACC) of learning from relavant keywords and unlabeled documents \textbf{with the threshold based on Algorithm~\ref{alg:pseudo}}. Outperforming methods are highlighted in boldface using one-sided t-test with the significance level of 5\%.}
\resizebox{\textwidth}{!}{\begin{tabular} {|c|c|c|c|c|c|c|c|c|c|c|}
\hline
\multirow{2}{*}{Dataset}&\multirow{2}{*}{Evaluation}&\multicolumn{2}{c|}{Proposed framework} & \multicolumn{3}{c|}{Text-feature baselines} & \multicolumn{2}{c|}{GloVe-feature baselines} & \multicolumn{2}{c|}{Zero-shot baselines} \\
&& Sigmoid& Logistic& PU-NB& NB& Maxent& RandomForest& KNN& GloveRanking& Voting\\ \hline
\multirow{2}{*}{$\mathsf{Subj}$}&$\mathrm{F}_{1}$ & \textbf{80.1 (0.35)} &76.0 (0.32) &47.1 (0.20) &76.9 (0.24) &76.8 (0.21) &54.8 (0.13) &61.2 (0.27) &74.3 (0.14) &63.5 (0.18)\\ 
&ACC & \textbf{80.2 (0.35)} &76.1 (0.32) &55.0 (0.13) &77.0 (0.24) &76.8 (0.21) &60.9 (0.09) &64.0 (0.21) &74.3 (0.14) &64.1 (0.18)\\ 
\hline 
\multirow{2}{*}{$\mathsf{Custrev}$}&$\mathrm{F}_{1}$ &62.3 (0.33) &61.8 (0.33) &58.4 (0.34) & \textbf{64.1 (0.32)} &60.8 (0.35) &55.6 (0.46) &60.3 (0.48) &50.9 (0.29) &61.9 (0.32)\\ 
&ACC &62.4 (0.33) &61.9 (0.33) & \textbf{64.3 (0.33)} & \textbf{64.3 (0.33)} &60.9 (0.36) &55.7 (0.46) &60.9 (0.59) &51.0 (0.30) &63.2 (0.34)\\ 
\hline 
\multirow{2}{*}{$\mathsf{MPQA}$}&$\mathrm{F}_{1}$ & \textbf{52.8 (0.26)} &52.1 (0.26) &46.7 (0.23) &49.0 (0.21) &51.4 (0.22) &52.2 (0.26) & \textbf{53.2 (0.29)} &47.6 (0.12) &53.3 (0.19)\\ 
&ACC & \textbf{71.9 (0.11)} &71.5 (0.11) &47.1 (0.24) &70.4 (0.07) &66.1 (0.13) &71.4 (0.12) &71.5 (0.09) &68.7 (0.07) &71.4 (0.07)\\ 
\hline 
\multirow{2}{*}{$\mathsf{AYI}$}&$\mathrm{F}_{1}$ & \textbf{64.5 (0.44)} & \textbf{64.1 (0.35)} &58.9 (0.47) &62.8 (0.49) &54.5 (0.46) &58.8 (0.52) & \textbf{64.9 (0.52)} &55.5 (0.45) &33.5 (0.00)\\ 
&ACC & \textbf{66.5 (0.38)} & \textbf{66.2 (0.31)} &60.1 (0.41) &65.1 (0.39) &57.1 (0.36) &61.4 (0.43) & \textbf{65.9 (0.43)} &58.4 (0.41) &50.5 (0.00)\\ 
\hline 
\multirow{2}{*}{$\mathsf{20NG}$}&$\mathrm{F}_{1}$ & \textbf{83.8 (0.24)} & \textbf{83.8 (0.23)} &N/A &58.4 (0.22) &51.8 (0.22) &83.2 (0.32) & \textbf{84.3 (0.26)} &83.1 (0.26) &9.6 (0.00)\\ 
&ACC & \textbf{95.1 (0.06)} & \textbf{95.1 (0.06)} &N/A &70.2 (0.31) &85.1 (0.09) &95.0 (0.08) & \textbf{95.1 (0.06)} &94.9 (0.07) &10.6 (0.00)\\ 
\hline 
\end{tabular}}
\label{tab:heu-thr}
\end{table}
\subsection{True Threshold $\pi$}
Table~\ref{tab:true-ref-thr} reports a result as we can see in the main body of the paper.

\begin{table}[htb]
\centering
\caption{Mean value and standard error of 20 trials for the $\mathrm{F}_{1}$-measure, and accuracy (ACC) of learning from relavant keywords and unlabeled documents \textbf{with the true threshold}. Outperforming methods are highlighted in boldface using one-sided t-test with the significance level of 5\%.}
\resizebox{\textwidth}{!}{\begin{tabular} {|c|c|c|c|c|c|c|c|c|c|c|c|c|}
\hline
\multirow{2}{*}{Dataset}&\multirow{2}{*}{Evaluation}&\multicolumn{2}{c|}{Proposed framework} & \multicolumn{3}{c|}{Text-feature baselines} & \multicolumn{2}{c|}{GloVe-feature baselines} & \multicolumn{2}{c|}{Zero-shot baselines} \\
&& Sigmoid& Logistic& PU-NB& NB& Maxent& Randomforest& Knn& GloveRanking& Voting\\ \hline
\multirow{2}{*}{$\mathsf{Subj}$}&$\mathrm{F}_{1}$ & \textbf{80.1 (0.38)} &76.0 (0.32) &47.1 (0.20) &76.3 (0.16) &76.3 (0.24) &75.1 (0.27) &63.6 (0.32) &74.5 (0.13) &63.5 (0.18)\\ 
&ACC & \textbf{80.1 (0.38)} &76.0 (0.32) &55.0 (0.13) &76.3 (0.16) &76.3 (0.24) &75.1 (0.27) &65.0 (0.28) &74.5 (0.13) &64.1 (0.18)\\ 
\hline 
\multirow{2}{*}{$\mathsf{Custrev}$}&$\mathrm{F}_{1}$ & \textbf{63.6 (0.41)} & \textbf{63.1 (0.31)} &58.4 (0.35) & \textbf{63.3 (0.49)} &62.1 (0.43) &58.3 (0.44) &60.8 (0.33) &53.0 (0.40) &38.9 (0.00)\\ 
&ACC &66.5 (0.40) &66.0 (0.31) &64.8 (0.32) & \textbf{69.8 (0.37)} &65.8 (0.39) &61.9 (0.40) &66.6 (0.29) &56.6 (0.37) &63.7 (0.00)\\ 
\hline 
\multirow{2}{*}{$\mathsf{MPQA}$}&$\mathrm{F}_{1}$ & \textbf{71.7 (0.44)} &69.8 (0.31) &46.7 (0.23) &54.3 (0.28) &53.1 (0.20) &62.4 (0.45) &23.8 (0.00) &57.5 (0.17) &23.8 (0.00)\\ 
&ACC & \textbf{75.6 (0.39)} &74.0 (0.27) &47.1 (0.24) &62.4 (0.28) &58.4 (0.17) &67.4 (0.39) &31.2 (0.00) &63.3 (0.17) &31.2 (0.00)\\ 
\hline 
\multirow{2}{*}{$\mathsf{AYI}$}&$\mathrm{F}_{1}$ & \textbf{69.3 (0.36)} & \textbf{68.8 (0.40)} &58.9 (0.47) &61.6 (0.38) &56.6 (0.36) &64.5 (0.53) &65.5 (0.52) &58.7 (0.51) &33.5 (0.00)\\ 
&ACC & \textbf{69.3 (0.36)} & \textbf{68.8 (0.40)} &60.1 (0.41) &62.5 (0.35) &56.8 (0.36) &64.6 (0.53) &65.8 (0.44) &58.7 (0.51) &50.5 (0.00)\\ 
\hline 
\multirow{2}{*}{$\mathsf{20NG}$}&$\mathrm{F}_{1}$ & \textbf{90.8 (0.20)} & \textbf{90.6 (0.21)} &N/A &58.4 (0.22) &52.4 (0.25) &89.6 (0.28) &86.7 (0.59) &90.5 (0.18) &9.6 (0.00)\\ 
&ACC & \textbf{96.5 (0.08)} & \textbf{96.4 (0.08)} &N/A &70.2 (0.31) &81.6 (0.11) &96.1 (0.10) &94.5 (0.35) &96.4 (0.07) &10.6 (0.00)\\ 
\hline 

\end{tabular}}
\label{tab:true-ref-thr}
\end{table}
\pagebreak
\subsection{Threshold with Varying $\hat{\pi}$}
Tables 10--14 illustrate the performance under varying $\hat{\pi}$ for five datasets.
\begin{table}[htb]
\centering
\caption{Mean $\mathrm{F}_{1}$-measure and standard error of 20 trials with varying thresholds with different $\hat{\pi}$ in Eq.~\eqref{eq:thr}: {$\mathsf{Subj}$}, $\pi=0.5$}
\resizebox{\textwidth}{!}{\begin{tabular} { |c|c|c|c|c|c|c|c|c|c|c|c|}
\hline
Evaluation & Methods& 0.05& 0.1& 0.2& 0.3& 0.4& 0.5& 0.6& 0.7& 0.8& 0.9\\ \hline
\multirow{9}{*}{$\mathrm{F}_{1}$}&Sigmoid &43.3 (0.17) &51.2 (0.24) &63.9 (0.34) &72.5 (0.27) &77.9 (0.37) & \textbf{80.1 (0.38)} &78.7 (0.32) &73.7 (0.29) &64.7 (0.32) &51.8 (0.30)\\ 
&Logistic &43.3 (0.17) &50.3 (0.25) &62.0 (0.28) &69.8 (0.26) &74.2 (0.29) & \textbf{76.0 (0.32)} &75.2 (0.28) &71.2 (0.31) &63.6 (0.32) &51.3 (0.34)\\ 
&PU-NB & \textbf{47.1 (0.20)} & \textbf{47.1 (0.20)} & \textbf{47.1 (0.20)} & \textbf{47.1 (0.20)} & \textbf{47.1 (0.20)} & \textbf{47.1 (0.20)} & \textbf{47.1 (0.20)} & \textbf{47.1 (0.20)} & \textbf{47.1 (0.20)} & \textbf{47.1 (0.20)}\\ 
&NB &37.9 (0.15) &42.3 (0.17) &50.7 (0.27) &60.0 (0.24) &69.0 (0.22) & \textbf{76.3 (0.16)} &72.0 (0.23) &59.1 (0.20) &46.7 (0.23) &37.7 (0.16)\\ 
&Maxent &37.1 (0.15) &41.9 (0.18) &52.5 (0.27) &62.1 (0.19) &70.6 (0.27) & \textbf{76.3 (0.24)} &74.5 (0.25) &64.1 (0.25) &50.3 (0.22) &39.3 (0.13)\\ 
&RandomForest &42.1 (0.22) &49.9 (0.15) &61.6 (0.22) &69.1 (0.26) &73.5 (0.29) & \textbf{75.1 (0.27)} &73.5 (0.17) &69.0 (0.21) &61.2 (0.24) &49.6 (0.23)\\ 
&KNN &43.6 (0.23) &51.7 (0.26) &61.2 (0.27) & \textbf{68.3 (0.27)} & \textbf{67.7 (0.48)} &63.6 (0.32) &63.6 (0.32) &34.7 (1.41) &33.3 (0.00) &33.3 (0.00)\\ 
&GloveRanking &43.1 (0.24) &50.5 (0.25) &61.9 (0.25) &69.3 (0.27) &73.6 (0.18) & \textbf{74.5 (0.13)} &72.8 (0.18) &68.0 (0.16) &60.6 (0.23) &49.4 (0.22)\\ 
&Voting &43.2 (0.16) &49.3 (0.17) &59.9 (0.21) &63.1 (0.22) & \textbf{65.2 (0.22)} &63.5 (0.18) &63.5 (0.18) &56.0 (0.20) &33.3 (0.00) &33.3 (0.00)\\ 
\hline 
\multirow{9}{*}{ACC}&Sigmoid &54.7 (0.09) &59.0 (0.15) &67.1 (0.27) &73.6 (0.26) &78.1 (0.36) & \textbf{80.1 (0.38)} &78.9 (0.31) &74.7 (0.26) &67.9 (0.24) &59.5 (0.18)\\ 
&Logistic &54.7 (0.09) &58.2 (0.15) &65.4 (0.21) &71.0 (0.23) &74.4 (0.28) & \textbf{76.0 (0.32)} & \textbf{75.4 (0.28)} &72.3 (0.28) &66.9 (0.25) &59.1 (0.20)\\ 
&PU-NB & \textbf{55.0 (0.13)} & \textbf{55.0 (0.13)} & \textbf{55.0 (0.13)} & \textbf{55.0 (0.13)} & \textbf{55.0 (0.13)} & \textbf{55.0 (0.13)} & \textbf{55.0 (0.13)} & \textbf{55.0 (0.13)} & \textbf{55.0 (0.13)} & \textbf{55.0 (0.13)}\\ 
&NB &52.1 (0.07) &54.2 (0.09) &58.5 (0.17) &64.1 (0.19) &70.4 (0.19) & \textbf{76.3 (0.16)} &73.3 (0.20) &64.0 (0.14) &56.5 (0.13) &51.9 (0.08)\\ 
&Maxent &51.7 (0.07) &54.0 (0.10) &59.6 (0.17) &65.4 (0.14) &71.5 (0.25) & \textbf{76.3 (0.24)} &75.1 (0.24) &67.4 (0.19) &58.6 (0.14) &52.7 (0.07)\\ 
&RandomForest &54.0 (0.12) &58.1 (0.10) &65.1 (0.18) &70.3 (0.24) &73.8 (0.29) & \textbf{75.1 (0.27)} &73.8 (0.17) &70.4 (0.19) &64.8 (0.18) &57.8 (0.14)\\ 
&KNN &54.7 (0.13) &58.7 (0.18) &64.0 (0.21) & \textbf{68.6 (0.26)} & \textbf{68.1 (0.42)} &65.0 (0.28) &65.0 (0.28) &50.7 (0.67) &50.00 (0.00) &50.00 (0.00)\\ 
&GloveRanking &54.6 (0.13) &58.4 (0.15) &65.4 (0.21) &70.6 (0.25) &73.8 (0.18) & \textbf{74.5 (0.13)} &73.0 (0.17) &69.3 (0.14) &64.1 (0.17) &57.5 (0.13)\\ 
&Voting &54.1 (0.10) &57.0 (0.12) &62.0 (0.19) &63.8 (0.21) & \textbf{65.2 (0.22)} &64.1 (0.18) &64.1 (0.18) &59.9 (0.15) &50.00 (0.00) &50.00 (0.00)\\ 
\hline 
\end{tabular}}
\end{table}

\begin{table}[htb]
\centering
\caption{Mean $\mathrm{F}_{1}$-measure and standard error of 20 trials with varying thresholds with different $\hat{\pi}$ in Eq.~\eqref{eq:thr} : {$\mathsf{Custrev}$}, $\pi=0.65$}
\resizebox{\textwidth}{!}{\begin{tabular} { |c|c|c|c|c|c|c|c|c|c|c|c|}
\hline
Evaluation & Methods& 0.05& 0.1& 0.2& 0.3& 0.4& 0.5& 0.6& 0.7& 0.8& 0.9\\ \hline
\multirow{9}{*}{$\mathrm{F}_{1}$}&Sigmoid &34.0 (0.24) &40.2 (0.33) &50.7 (0.43) &57.6 (0.36) &62.1 (0.37) & \textbf{63.9 (0.37)} & \textbf{64.0 (0.38)} &62.8 (0.33) &58.3 (0.40) &51.7 (0.34)\\ 
&Logistic &34.2 (0.26) &40.4 (0.42) &51.1 (0.42) &58.0 (0.31) &61.6 (0.35) & \textbf{62.9 (0.31)} & \textbf{63.2 (0.34)} &62.1 (0.43) &58.2 (0.33) &50.8 (0.36)\\ 
&PU-NB & \textbf{58.4 (0.34)} & \textbf{58.4 (0.34)} & \textbf{58.4 (0.34)} & \textbf{58.4 (0.34)} & \textbf{58.4 (0.34)} & \textbf{58.4 (0.34)} & \textbf{58.4 (0.34)} & \textbf{57.9 (0.40)} &53.3 (0.32) &46.1 (0.26)\\ 
&NB &28.9 (0.17) &32.6 (0.20) &40.4 (0.36) &49.5 (0.45) &62.8 (0.43) & \textbf{67.1 (0.41)} &64.7 (0.48) &60.2 (0.46) &53.4 (0.39) &45.9 (0.37)\\ 
&Maxent &31.1 (0.22) &36.3 (0.21) &45.4 (0.32) &53.8 (0.39) &60.1 (0.35) & \textbf{62.6 (0.48)} & \textbf{62.7 (0.48)} &60.7 (0.49) &56.1 (0.52) &47.9 (0.46)\\ 
&RandomForest &31.9 (0.23) &36.5 (0.18) &45.0 (0.28) &50.8 (0.36) &55.3 (0.47) & \textbf{57.7 (0.43)} & \textbf{58.5 (0.40)} & \textbf{57.5 (0.44)} &54.0 (0.35) &47.7 (0.33)\\ 
&KNN &38.3 (0.33) &46.8 (0.35) &54.8 (0.30) &58.9 (0.63) &60.1 (0.44) & \textbf{61.9 (0.38)} &60.8 (0.33) &60.5 (0.33) &55.1 (0.30) &47.4 (0.29)\\ 
&GloveRanking &32.8 (0.31) &37.2 (0.29) &44.0 (0.29) &47.6 (0.22) &50.6 (0.27) &52.0 (0.30) & \textbf{53.1 (0.45)} & \textbf{53.1 (0.41)} &50.4 (0.31) &45.9 (0.24)\\ 
&Voting &36.0 (0.27) &41.3 (0.50) &52.9 (0.34) &60.6 (0.33) &61.1 (0.28) & \textbf{62.0 (0.34)} &38.9 (0.00) &38.9 (0.00) &38.9 (0.00) &38.9 (0.00)\\ 
\hline 
\multirow{9}{*}{ACC}&Sigmoid &40.5 (0.15) &44.4 (0.23) &52.0 (0.36) &57.8 (0.34) &62.1 (0.38) &64.6 (0.39) &66.1 (0.39) & \textbf{67.3 (0.34)} & \textbf{66.6 (0.29)} &65.8 (0.20)\\ 
&Logistic &40.7 (0.16) &44.6 (0.29) &52.5 (0.36) &58.2 (0.29) &61.7 (0.35) &63.6 (0.34) &65.4 (0.34) & \textbf{66.7 (0.39)} & \textbf{66.6 (0.24)} &65.3 (0.20)\\ 
&PU-NB &64.3 (0.33) &64.3 (0.33) &64.3 (0.33) &64.3 (0.33) &64.3 (0.33) &64.3 (0.33) &64.3 (0.33) & \textbf{66.9 (0.28)} & \textbf{67.3 (0.17)} &65.7 (0.12)\\ 
&NB &37.6 (0.10) &39.7 (0.13) &44.5 (0.26) &51.0 (0.36) &62.9 (0.44) & \textbf{69.2 (0.38)} & \textbf{69.9 (0.38)} & \textbf{69.5 (0.31)} &67.7 (0.21) &65.8 (0.15)\\ 
&Maxent &38.8 (0.14) &42.0 (0.14) &48.0 (0.25) &54.4 (0.36) &60.1 (0.35) &63.4 (0.48) &65.3 (0.42) & \textbf{66.5 (0.38)} & \textbf{66.6 (0.35)} &65.3 (0.25)\\ 
&RandomForest &39.1 (0.14) &41.6 (0.11) &46.9 (0.25) &51.2 (0.35) &55.4 (0.47) &58.6 (0.43) &61.2 (0.39) &63.0 (0.35) & \textbf{63.9 (0.20)} & \textbf{64.0 (0.16)}\\ 
&KNN &43.0 (0.22) &48.7 (0.29) &55.0 (0.28) &59.2 (0.63) &60.4 (0.51) &64.1 (0.39) & \textbf{66.6 (0.29)} & \textbf{66.6 (0.27)} & \textbf{66.4 (0.20)} &65.2 (0.13)\\ 
&GloveRanking &39.4 (0.20) &41.6 (0.22) &45.6 (0.25) &47.8 (0.20) &50.7 (0.28) &52.9 (0.31) &55.7 (0.43) &58.4 (0.34) &59.9 (0.30) & \textbf{61.5 (0.17)}\\ 
&Voting &41.5 (0.19) &44.9 (0.35) &53.5 (0.30) &60.6 (0.33) &61.3 (0.32) & \textbf{63.3 (0.34)} & \textbf{63.7 (0.00)} & \textbf{63.7 (0.00)} & \textbf{63.7 (0.00)} & \textbf{63.7 (0.00)}\\ 
\hline 
\end{tabular}}
\end{table}

\begin{table}[htb]
\centering
\caption{Mean $\mathrm{F}_{1}$-measure and standard error of 20 trials with varying thresholds with different $\hat{\pi}$ in Eq.~\eqref{eq:thr} : {$\mathsf{MPQA}$}, $\pi=0.31$}
\resizebox{\textwidth}{!}{\begin{tabular} { |c|c|c|c|c|c|c|c|c|c|c|c|}
\hline
Evaluation & Methods& 0.05& 0.1& 0.2& 0.3& 0.4& 0.5& 0.6& 0.7& 0.8& 0.9\\ \hline
\multirow{9}{*}{$\mathrm{F}_{1}$}&Sigmoid &52.5 (0.28) &60.7 (0.38) &68.7 (0.43) & \textbf{71.6 (0.45)} & \textbf{71.0 (0.41)} &68.3 (0.36) &63.4 (0.34) &56.6 (0.29) &48.6 (0.18) &37.6 (0.25)\\ 
&Logistic &52.0 (0.27) &59.0 (0.25) &67.1 (0.33) & \textbf{69.8 (0.31)} & \textbf{69.5 (0.37)} &67.4 (0.33) &62.9 (0.29) &56.3 (0.25) &48.1 (0.25) &37.6 (0.25)\\ 
&PU-NB & \textbf{46.7 (0.23)} & \textbf{46.7 (0.23)} & \textbf{46.7 (0.23)} & \textbf{46.7 (0.23)} & \textbf{46.7 (0.23)} & \textbf{46.7 (0.23)} & \textbf{46.7 (0.23)} & \textbf{46.6 (0.23)} &42.9 (0.24) &34.2 (0.22)\\ 
&NB &48.9 (0.23) &52.7 (0.22) & \textbf{54.1 (0.23)} & \textbf{54.3 (0.26)} & \textbf{53.8 (0.26)} &52.6 (0.25) &50.2 (0.24) &47.6 (0.22) &40.5 (0.21) &34.5 (0.23)\\ 
&Maxent &51.4 (0.22) & \textbf{52.9 (0.18)} & \textbf{52.8 (0.22)} & \textbf{53.1 (0.20)} &51.9 (0.20) &51.9 (0.20) &49.5 (0.16) &45.9 (0.20) &40.2 (0.18) &33.4 (0.17)\\ 
&RandomForest &52.0 (0.28) &56.6 (0.36) &61.1 (0.43) & \textbf{62.3 (0.44)} & \textbf{61.7 (0.45)} &60.0 (0.38) &56.5 (0.38) &51.3 (0.33) &44.6 (0.34) &35.9 (0.19)\\ 
&KNN &52.8 (0.24) &56.7 (0.26) & \textbf{59.4 (0.26)} &23.8 (0.00) &23.8 (0.00) &23.8 (0.00) &23.8 (0.00) &23.8 (0.00) &23.8 (0.00) &23.8 (0.00)\\ 
&GloveRanking &47.5 (0.12) &50.9 (0.19) &55.5 (0.21) & \textbf{57.5 (0.19)} & \textbf{57.9 (0.18)} &57.0 (0.25) &54.9 (0.21) &51.0 (0.19) &44.2 (0.23) &35.7 (0.20)\\ 
&Voting & \textbf{53.3 (0.19)} &23.8 (0.00) &23.8 (0.00) &23.8 (0.00) &23.8 (0.00) &23.8 (0.00) &23.8 (0.00) &23.8 (0.00) &23.8 (0.00) &23.8 (0.00)\\ 
\hline 
\multirow{9}{*}{ACC}&Sigmoid &71.9 (0.12) &74.2 (0.21) & \textbf{76.1 (0.32)} & \textbf{75.8 (0.40)} &73.4 (0.38) &69.4 (0.35) &63.7 (0.35) &56.6 (0.29) &49.2 (0.16) &40.4 (0.19)\\ 
&Logistic &71.4 (0.10) &73.1 (0.18) & \textbf{74.8 (0.26)} & \textbf{74.3 (0.27)} &72.0 (0.35) &68.5 (0.33) &63.1 (0.29) &56.3 (0.25) &48.8 (0.23) &40.4 (0.19)\\ 
&PU-NB & \textbf{47.1 (0.24)} & \textbf{47.1 (0.24)} & \textbf{47.1 (0.24)} & \textbf{47.1 (0.24)} & \textbf{47.1 (0.24)} & \textbf{47.1 (0.24)} & \textbf{47.1 (0.24)} & \textbf{46.8 (0.22)} &43.0 (0.23) &37.0 (0.18)\\ 
&NB & \textbf{70.4 (0.08)} &70.1 (0.11) &67.0 (0.16) &63.4 (0.22) &59.7 (0.24) &56.0 (0.23) &51.4 (0.25) &47.8 (0.22) &41.1 (0.20) &37.4 (0.18)\\ 
&Maxent & \textbf{66.2 (0.13)} &64.0 (0.16) &60.2 (0.17) &58.6 (0.17) &55.6 (0.16) &53.2 (0.20) &49.7 (0.16) &46.0 (0.20) &41.1 (0.16) &36.6 (0.14)\\ 
&RandomForest & \textbf{71.4 (0.12)} & \textbf{71.5 (0.20)} &70.2 (0.30) &67.7 (0.38) &64.7 (0.42) &61.3 (0.38) &56.8 (0.38) &51.3 (0.33) &45.4 (0.32) &38.9 (0.16)\\ 
&KNN & \textbf{71.5 (0.08)} &70.7 (0.21) &67.4 (0.23) &31.2 (0.00) &31.2 (0.00) &31.2 (0.00) &31.2 (0.00) &31.2 (0.00) &31.2 (0.00) &31.2 (0.00)\\ 
&GloveRanking & \textbf{68.7 (0.06)} &67.6 (0.11) &65.9 (0.17) &63.8 (0.18) &61.4 (0.17) &58.5 (0.25) &55.2 (0.22) &51.0 (0.19) &44.9 (0.21) &38.6 (0.16)\\ 
&Voting & \textbf{71.4 (0.07)} &31.2 (0.00) &31.2 (0.00) &31.2 (0.00) &31.2 (0.00) &31.2 (0.00) &31.2 (0.00) &31.2 (0.00) &31.2 (0.00) &31.2 (0.00)\\ 
\hline 
\end{tabular}}
\end{table}

\begin{table}[htb]
\centering
\caption{Mean $\mathrm{F}_{1}$-measure and standard error of 20 trials with varying thresholds with different $\hat{\pi}$ in Eq.~\eqref{eq:thr} : {$\mathsf{AYI}$}, $\pi=0.50$}
\resizebox{\textwidth}{!}{\begin{tabular} { |c|c|c|c|c|c|c|c|c|c|c|c|}
\hline
Evaluation & Methods& 0.05& 0.1& 0.2& 0.3& 0.4& 0.5& 0.6& 0.7& 0.8& 0.9\\ \hline
\multirow{9}{*}{$\mathrm{F}_{1}$}&Sigmoid &42.8 (0.44) &51.3 (0.50) &61.2 (0.44) &66.5 (0.44) & \textbf{68.6 (0.42)} & \textbf{69.3 (0.36)} &66.9 (0.41) &63.5 (0.46) &56.8 (0.52) &47.9 (0.35)\\ 
&Logistic &43.1 (0.49) &50.6 (0.48) &60.9 (0.43) &66.5 (0.40) & \textbf{68.7 (0.43)} & \textbf{68.8 (0.40)} &66.6 (0.42) &62.7 (0.42) &56.4 (0.54) &46.9 (0.36)\\ 
&PU-NB & \textbf{58.9 (0.47)} & \textbf{58.9 (0.47)} & \textbf{58.9 (0.47)} & \textbf{58.9 (0.47)} & \textbf{58.9 (0.47)} & \textbf{58.9 (0.47)} & \textbf{58.2 (0.38)} &53.8 (0.38) &48.0 (0.36) &40.5 (0.38)\\ 
&NB &33.7 (0.12) &36.1 (0.20) &49.9 (0.44) & \textbf{65.9 (0.47)} & \textbf{64.8 (0.44)} &61.7 (0.38) &57.5 (0.47) &52.3 (0.42) &46.5 (0.38) &39.2 (0.41)\\ 
&Maxent &39.0 (0.32) &45.4 (0.48) &52.6 (0.52) &56.0 (0.43) & \textbf{57.1 (0.41)} & \textbf{56.8 (0.36)} &54.3 (0.41) &50.9 (0.41) &45.8 (0.38) &38.8 (0.32)\\ 
&RandomForest &39.6 (0.47) &46.4 (0.42) &55.3 (0.55) &61.4 (0.52) & \textbf{64.5 (0.51)} & \textbf{64.5 (0.54)} &62.8 (0.47) &58.1 (0.60) &52.8 (0.57) &44.1 (0.42)\\ 
&KNN &44.5 (0.33) &50.2 (0.58) &61.6 (0.43) & \textbf{65.6 (0.49)} & \textbf{66.5 (0.41)} & \textbf{65.6 (0.50)} &61.1 (0.35) &54.4 (1.23) &49.9 (0.34) &33.5 (0.00)\\ 
&GloveRanking &39.9 (0.39) &46.1 (0.56) &53.7 (0.46) &56.6 (0.55) & \textbf{58.0 (0.59)} & \textbf{58.7 (0.51)} & \textbf{57.8 (0.40)} &55.2 (0.41) &50.2 (0.30) &43.3 (0.36)\\ 
&Voting &43.7 (0.31) & \textbf{52.4 (0.38)} &46.6 (2.47) &33.5 (0.00) &33.5 (0.00) &33.5 (0.00) &33.5 (0.00) &33.5 (0.00) &33.5 (0.00) &33.5 (0.00)\\ 
\hline 
\multirow{9}{*}{ACC}&Sigmoid &54.1 (0.24) &58.7 (0.32) &64.4 (0.32) &67.8 (0.39) & \textbf{68.9 (0.41)} & \textbf{69.3 (0.36)} &67.4 (0.38) &65.1 (0.39) &61.0 (0.38) &56.4 (0.23)\\ 
&Logistic &54.4 (0.26) &58.2 (0.30) &64.2 (0.33) &67.7 (0.38) & \textbf{69.0 (0.42)} & \textbf{68.8 (0.40)} &67.1 (0.40) &64.4 (0.36) &60.7 (0.40) &55.9 (0.22)\\ 
&PU-NB & \textbf{60.1 (0.41)} & \textbf{60.1 (0.41)} & \textbf{60.1 (0.41)} & \textbf{60.1 (0.41)} & \textbf{60.1 (0.41)} & \textbf{60.1 (0.41)} & \textbf{59.8 (0.36)} &57.8 (0.28) &55.4 (0.22) &52.6 (0.19)\\ 
&NB &49.7 (0.06) &50.8 (0.10) &57.6 (0.25) & \textbf{66.4 (0.44)} &64.9 (0.44) &62.5 (0.36) &60.0 (0.40) &57.4 (0.30) &55.0 (0.23) &52.2 (0.22)\\ 
&Maxent &52.3 (0.16) &54.7 (0.29) & \textbf{56.9 (0.39)} & \textbf{57.4 (0.38)} & \textbf{57.2 (0.40)} & \textbf{56.9 (0.36)} &55.6 (0.37) &54.5 (0.34) &53.2 (0.24) &51.6 (0.19)\\ 
&RandomForest &52.4 (0.24) &55.5 (0.28) &59.6 (0.43) &63.0 (0.46) & \textbf{64.8 (0.48)} & \textbf{64.6 (0.54)} &63.5 (0.45) &60.6 (0.54) &58.2 (0.44) &54.2 (0.29)\\ 
&KNN &54.9 (0.19) &57.8 (0.35) &64.1 (0.34) & \textbf{66.1 (0.47)} & \textbf{66.6 (0.40)} & \textbf{65.9 (0.43)} &62.9 (0.31) &59.3 (0.68) &57.0 (0.23) &50.5 (0.00)\\ 
&GloveRanking &52.0 (0.18) &54.8 (0.30) & \textbf{58.0 (0.36)} & \textbf{58.5 (0.52)} & \textbf{58.6 (0.57)} & \textbf{58.7 (0.51)} & \textbf{58.2 (0.41)} &57.0 (0.38) &54.7 (0.25) &52.6 (0.22)\\ 
&Voting &53.8 (0.17) & \textbf{58.8 (0.25)} &55.9 (1.04) &50.5 (0.00) &50.5 (0.00) &50.5 (0.00) &50.5 (0.00) &50.5 (0.00) &50.5 (0.00) &50.5 (0.00)\\ 
\hline 
\end{tabular}}
\end{table}

\begin{table}[htb]
\centering
\caption{Mean $\mathrm{F}_{1}$-measure and standard error of 20 trials with varying thresholds with different $\hat{\pi}$ in Eq.~\eqref{eq:thr} : {$\mathsf{20NG}$}, true $\pi=0.11$}
\resizebox{\textwidth}{!}{\begin{tabular} { |c|c|c|c|c|c|c|c|c|c|c|c|}
\hline
Evaluation & Methods& 0.05& 0.1& 0.2& 0.3& 0.4& 0.5& 0.6& 0.7& 0.8& 0.9\\ \hline
\multirow{8}{*}{$\mathrm{F}_{1}$}&Sigmoid &79.9 (0.23) & \textbf{91.2 (0.18)} &78.4 (0.20) &68.2 (0.15) &60.0 (0.14) &52.7 (0.13) &45.3 (0.15) &37.8 (0.13) &29.7 (0.13) &20.5 (0.13)\\ 
&Logistic &79.7 (0.29) & \textbf{90.9 (0.18)} &78.2 (0.20) &68.0 (0.14) &59.9 (0.15) &52.6 (0.13) &45.4 (0.15) &37.9 (0.15) &29.8 (0.10) &20.6 (0.09)\\ 
&NB & \textbf{58.4 (0.22)} & \textbf{58.4 (0.22)} & \textbf{58.4 (0.22)} & \textbf{58.4 (0.22)} & \textbf{58.4 (0.22)} &54.4 (0.14) &48.3 (0.10) &41.4 (0.10) &33.6 (0.12) &24.4 (0.12)\\ 
&Maxent &51.5 (0.19) & \textbf{52.3 (0.24)} &51.5 (0.21) &49.5 (0.16) &46.6 (0.15) &43.0 (0.16) &38.1 (0.14) &32.9 (0.15) &26.3 (0.16) &19.0 (0.13)\\ 
&RandomForest &79.4 (0.33) & \textbf{89.8 (0.29)} &78.5 (0.17) &68.2 (0.15) &59.8 (0.16) &52.6 (0.13) &45.4 (0.12) &37.7 (0.16) &29.6 (0.13) &20.3 (0.10)\\ 
&KNN &83.9 (0.41) & \textbf{88.4 (0.19)} &68.4 (0.20) &9.6 (0.00) &9.6 (0.00) &9.6 (0.00) &9.6 (0.00) &9.6 (0.00) &9.6 (0.00) &9.6 (0.00)\\ 
&GloveRanking &79.2 (0.34) & \textbf{90.7 (0.17)} &78.2 (0.22) &67.9 (0.19) &59.9 (0.14) &52.3 (0.11) &45.0 (0.10) &37.6 (0.10) &29.3 (0.11) &20.1 (0.13)\\ 
&Voting & \textbf{9.6 (0.00)} & \textbf{9.6 (0.00)} & \textbf{9.6 (0.00)} & \textbf{9.6 (0.00)} & \textbf{9.6 (0.00)} & \textbf{9.6 (0.00)} & \textbf{9.6 (0.00)} & \textbf{9.6 (0.00)} & \textbf{9.6 (0.00)} & \textbf{9.6 (0.00)}\\ 
\hline 
\multirow{8}{*}{ACC}&Sigmoid &94.3 (0.05) & \textbf{96.8 (0.07)} &88.8 (0.14) &79.4 (0.14) &69.7 (0.17) &60.1 (0.17) &49.9 (0.21) &40.0 (0.16) &30.3 (0.14) &20.5 (0.13)\\ 
&Logistic &94.2 (0.06) & \textbf{96.6 (0.07)} &88.7 (0.14) &79.3 (0.13) &69.7 (0.18) &60.0 (0.17) &50.1 (0.19) &40.2 (0.19) &30.4 (0.11) &20.6 (0.09)\\ 
&NB & \textbf{70.2 (0.31)} & \textbf{70.2 (0.31)} & \textbf{70.2 (0.31)} & \textbf{70.2 (0.31)} & \textbf{70.2 (0.31)} &64.0 (0.18) &55.0 (0.14) &45.2 (0.14) &35.1 (0.15) &24.5 (0.13)\\ 
&Maxent & \textbf{85.8 (0.08)} &82.0 (0.10) &74.4 (0.13) &66.8 (0.18) &59.1 (0.19) &51.3 (0.20) &43.0 (0.17) &35.2 (0.18) &26.8 (0.18) &19.0 (0.13)\\ 
&RandomForest &94.1 (0.07) & \textbf{96.3 (0.11)} &88.7 (0.12) &79.3 (0.13) &69.4 (0.19) &59.7 (0.17) &50.1 (0.16) &40.0 (0.20) &30.2 (0.14) &20.3 (0.10)\\ 
&KNN &95.0 (0.09) & \textbf{95.6 (0.08)} &79.7 (0.20) &10.6 (0.00) &10.6 (0.00) &10.6 (0.00) &10.6 (0.00) &10.6 (0.00) &10.6 (0.00) &10.6 (0.00)\\ 
&GloveRanking &94.0 (0.07) & \textbf{96.6 (0.06)} &88.7 (0.16) &79.3 (0.18) &69.7 (0.17) &59.7 (0.14) &49.7 (0.13) &40.0 (0.13) &29.9 (0.13) &20.1 (0.12)\\ 
&Voting & \textbf{10.6 (0.00)} & \textbf{10.6 (0.00)} & \textbf{10.6 (0.00)} & \textbf{10.6 (0.00)} & \textbf{10.6 (0.00)} & \textbf{10.6 (0.00)} & \textbf{10.6 (0.00)} & \textbf{10.6 (0.00)} & \textbf{10.6 (0.00)} & \textbf{10.6 (0.00)}\\ 
\hline 
\end{tabular}}
\end{table}

\end{document}